\documentstyle[fullpage,11pt,named,lingmacros]{article}  
\author{Marilyn A. Walker \\
ATT Laboratories -  Research \\
600 Mountain Ave. \\
Murray Hill, N.J. 07974 \\
{\tt walker@research.att.com}}

\input{psfig}

\title{Inferring Acceptance and Rejection in Dialogue \\ by Default Rules of Inference \\
\begin{small}  Language and Speech, 39-2, 1996; cmp-lg/9609002 \end{small}}

\date{}

\begin{document}          

\maketitle

\begin{abstract}
\begin{quote}
This paper discusses the processes by which conversants in a dialogue can infer
whether their assertions and proposals have been accepted or rejected by their
conversational partners. It expands on previous work by showing that logical
consistency is a necessary indicator of acceptance, but that it is not
sufficient, and that logical inconsistency is sufficient as an indicator of
rejection, but it is not necessary.  I show how conversants can use information
structure and prosody as well as logical reasoning in distinguishing between
acceptances and logically consistent rejections, and relate this work to previous
work on implicature and default reasoning by introducing three new classes of
rejection: {\sc implicature rejections}, {\sc epistemic rejections} and {\sc
deliberation rejections}. I show how these rejections are inferred as a result of
default inferences, which, by other analyses, would have been blocked by the
context. In order to account for these facts, I propose a model of the common
ground that allows these default inferences to go through, and show how the
model, originally proposed to account for the various forms of acceptance, can
also model all types of rejection.

\end{quote}
\end{abstract}


\bibliographystyle{named}


\section{Introduction}
\label{intro-sec}

A common view of dialogue is that the conversational record is part of
the {\sc common ground} of the conversants.  As conversants A and B
participate in a dialogue, A and B communicate with discourse acts
such as {\sc proposals, assertions, acceptances} and {\sc rejections}.
If A proposes an act $\alpha$ and B accepts A's proposal then the act
$\alpha$ becomes a mutual intention in A and B's common ground. If A
asserts a proposition $\phi$ and B accepts A's assertion, the $\phi$
becomes a mutual belief in the common ground. If B rejects A's
assertion or proposal, then what is rejected is not added to the
common ground \cite{Stalnaker78}.  Whenever conversants A and B must
remain coordinated on what is in the common ground \cite{Thomason90b},
they must monitor whether their utterances are accepted or rejected by
their conversational partners.

Coordinating on what is in the common ground would be straightforward
if conversants always explicitly indicated acceptance or rejection with 
forms such as {\it I agree} or {\it I reject your proposal}. However there are
a variety of ways to indicate both acceptance and rejection in natural dialogue,
most of which rely on making inferences from form to function.

For example, previous work suggests that B can indicate acceptance by
simply going on to the next topic, with phrases such as {\it uh huh,
okay, sure, alright}, or with full or partial repetitions.  For
example, consider the partial repetition in \ex{1}, by which B accepts
A's assertion.

\eenumsentence{
\item[A:] Sue's house is on Chestnut St.  
\item[B:] on Chestnut St.
\label{num-examp}
}

B can also indicate acceptance with paraphrases of what A said, and with
utterances in which B makes an inference explicit that follows from
what A has just asserted or proposed
\cite{Carberry89,CS89,WS88,Walker92a}.

It has been suggested that what unifies the forms of acceptance is
logical consistency with the context: explicit indicators of
acceptance provide `no new information' \cite{WS88}. Thus it would
appear that if A can identify the beginning of a new topic and detect
whether B's response is logically consistent with what A has said,
then A can infer whether B has accepted A's assertion or proposal.

The correlation between form and function for indicators of rejection
also appears to require A to make logical inferences. Horn's review of
the literature suggests that the types of rejection include: (a) {\sc
denial} as in \ex{1}; (b) {\sc logical contradiction} as in \ex{2};
 (c) {\sc implicit denial} as in \ex{3}, where B denies a
presupposition of A's; and (d) {\sc refusal}, also called {\sc
rejection} where B refuses an offer or proposal of A's
\cite{Horn89}.

\eenumsentence{
\item[A:] Pigs can fly.
\item[B:] No, you idiot, pigs can't fly! (Horn's 29)
\label{pig-examp}
}

\eenumsentence{
\item[A:] Kim and Lee have been partners since 1989.
\item[B:] But Lee said they met in 1990.
\label{kim-examp}
}

\eenumsentence{
\item[A:] Julia's daughter is a genius.
\item[B:] Julia doesn't have any children.
\label{julia-examp}
}

\eenumsentence{
\item[A:] Come and play ball with me.
\item[B:] No, I don't want to. (Horn's 33)
\label{ball-examp}
}

The assumption seems to have been that A can identify these forms as rejections
by recognizing logical inconsistency either directly from what was said, or via
an inferential chain \cite{Gazdar79,Allwood92} {\it inter alia}. Thus much of
previous work would lead one to believe that there is a neat trichotomy with the
processing that A must do to monitor additions to the common ground:

\begin{enumerate}
\item A detects logical consistency: leads A to an inference of {\sc explicit 
acceptance} 
\item A detects the initiation of a new topic: leads A to an inference of {\sc 
implicit acceptance} 
\item A detects logical inconsistency: leads A to an inference of {\sc 
rejection} 
\end{enumerate}

Unfortunately things are not so simple. 
Consider
the rejection in \ex{1}B:

\eenumsentence
{\item[A:] There's a man in the garage.  
\item[B:] There's something in the garage.
\label{garage-examp}
}

The proposition realized by \ex{0}B follows from \ex{0}A as an entailment via
existential generalization.  Thus \ex{0}B {\sc rejects} \ex{0}A while being
logically consistent with it.\footnote{The model of the common ground in section
\ref{lewis-sec} represents the fact that B's utterance is a partial acceptance as
well as a rejection, since it accepts a proposition entailed by A's assertion,
namely what B asserts in \ex{0}B.}  The only possible conclusion is that
rejections need not be logically inconsistent.  But if a logically consistent
utterance such as \ex{0}B can function as a rejection, how does A identify
\ex{0}B as a rejection? And how does A distinguish between logically consistent
utterances that indicate acceptance such as \ref{num-examp}B and utterances that
indicate rejection such as \ex{0}B? I argue that the inference of rejection
arises as a {\sc quantity implicature} from the fact that B's utterance was less
informative than it might have been \cite{Hirschberg85}, leading to the
implicature in \ex{1}:

\enumsentence{It's not a man. \label{garage-implic}}

However, this analysis raise several additional issues.  First, it is
well known that implicatures are default inferences that are
cancellable by prior context. Since the implicature in \ex{0} is not
consistent with the assertion in \ex{-1}A, how does the implicature
arise? Furthermore, if \ex{-1}B can implicate \ex{0} by only
confirming a part of what was asserted, why doesn't \ref{num-examp}B
potentially implicate \ex{1}?

\enumsentence{
It's not Sue's house.
\label{num-examp-implic}
}

The goal of this paper is to specify in more detail the processes by
which conversants infer whether their assertions and proposals have
been rejected by their conversational partners, and to specify what
features of the utterance they use in doing so. 

The discussion above suggests that logical consistency is necessary for
an indicator of explicit acceptance, but it is not sufficient. On the
other hand, logical inconsistency is sufficient as an indicator of
rejection, but it is not necessary.

Throughout the paper, I will ignore the existence of explicit forms for indicating
acceptance or rejection such as {\it Yes, I agree} or {\it No, I don't want to},
on the basis that their analysis is straightforward. In particular, nothing I
have to say about the linguistic cues for distinguishing acceptance from
rejection apply to utterances which explicitly assert acceptance or rejection.

In order to provide an empirical basis for this study, I collected 43 examples of
acceptance and 31 examples of rejection from a corpus of financial advice
dialogues and from the literature. \footnote{Most of the examples are from a
corpus of radio-talk show dialogues about financial advice, which consists of 55
dialogues from 5 hours of live broadcast. This corpus was first taped and
transcribed by Julia Hirschberg and Martha Pollack, and reported on in
\cite{PHW82}. There are many more examples of acceptance in the corpus than are
discussed here, but only 31 examples of rejection were identified in the 55
dialogues. I am grateful to Julia Hirschberg for providing me with audio tapes of
the original broadcast, and to Mark Liberman for prosodic analysis tools.}
Section \ref{acc-sec} discusses the forms for indicating acceptance and the cues
that conversants might use to infer acceptance.  Section \ref{reject-sec}
discusses the inference of rejection.  I use the corpus analysis to show that
there are many more types of rejection than previously noted, and that
conversants use information structure as well as logical reasoning in inferring
rejection. In section \ref{focus-sec}, I discuss features of utterances that are
useful in distinguishing acceptance from rejection, namely information structure
and prosodic realization.  In section \ref{lewis-sec}, I show that modeling the
function of these utterances requires a model of the common ground in which
beliefs can be only weakly mutual, and in which beliefs inferred by default rules
of inference are represented differently than beliefs inferred as entailments.

\section{Inferring Acceptance in Dialogue}
\label{acc-sec}

As mentioned above, there is a large amount of variation in the forms by which acceptance
is indicated \cite{CS89,WS88,WW90}.  This section first briefly discusses some features
of the dialogue model that underlie the analysis of acceptance presented below, and then
discusses the various forms of acceptance found in 43 examples from the financial advice
dialogue corpus \cite{PHW82} and in the literature. The main focus of this discussion is
the way in which the realization of these forms allows conversants to distinguish
acceptance from rejection, and thus stay coordinated on what is added to the common
ground during a conversation. I focus on forms which do not explicitly assert acceptance
since treatment of these should be straightforward.

A primary open issue about the various forms that indicate acceptance, is whether any
forms that don't explicitly assert {\it I believe you} or {\it I agree} indicate anything
more than simply understanding.  For example, a point of controversy is whether a phrase
such as {\it uh huh} can be used to indicate the adoption of belief or the commitment to
a proposed course of action \cite{Schegloff82,GS86}. I will argue that utterances that
add no new information, which occur as B's response, assert understanding with various
strengths, but only implicate acceptance.\footnote{Forms that add no new information
include prompts such as {\it uh huh}, repetitions, paraphrases, and utterances that make
inferences explicit.}

A secondary issue is that previous work has proposed that there is a class of {\sc
implicit acceptances}. The claim is that speaker A is licensed in inferring acceptance if
speaker B produces an utterance about a new topic in response to A's assertion or
proposal \cite{CS89}, or if B's response ``cannot be interpreted as initiating a
negotiation dialogue '' \cite{LambertCarberry91}. I will assume that under certain
circumstances, acceptance can be indicated implicitly, and will argue that the inferences
involved in inferring acceptance under these circumstances are the same as those involved
in making the inference from an assertion of understanding to acceptance, as discussed
above for explicit forms of acceptance.

Finally, I claim that both the `no new information' forms, and the cases of {\sc
implicit acceptance} only implicate acceptance under certain circumstances.

The first condition for implicating acceptance is that there is a conventional assumption
shared between the conversants of what I call the {\sc attitude locus}. The {\sc attitude
locus} is the sequential position in conversation, just after A's assertion or proposal,
in which B first has an opportunity to express B's attitude or evaluation of A's
utterance. I will show below that there is good evidence for the existence of the {\sc
attitude locus}.

The second condition, is that given the {\sc attitude locus}, conversants in particular
types of conversations, such as task-oriented or problem-solving dialogues, appear to
observe the rule of collaborative interaction in \ex{1}:

\enumsentence{
{\sc collaborative principle}: Conversants must provide
evidence of a detected discrepancy in belief as soon as
possible.
}

The {\sc collaborative principle} is an abstraction of the collaborative planning
principles proposed by \cite{WS88,WW90}.  The effect of this principle is that
conversants can make default inferences of acceptance from the fact that B has provided
no evidence of rejection or evidence that there is a need for clarification in the
attitude locus.

Thus, only the existence of the attitude locus and the assumption that conversants are
following the collaborative principle can license the implicature of acceptance from
forms that only indicate understanding, or from the implicit acceptance situations that
Clark and Carberry discuss.  Below I will refer to forms that only assert understanding,
but implicate acceptance, as forms of explicit acceptance.

Section \ref{form-acc-sec} discusses forms of explicit acceptance that provide no new
information. In each example, the explicit acceptance form is indicated in {\bf
boldface}, while the assertion or proposal that it accepts is indicated in {\it
italics}. In section \ref{lewis-sec}, I will show how the inference from these forms to
the implicature of acceptance is reflected by a model of the common ground in which the
implicated acceptances are defeasible.

\subsection{Forms of Explicit Acceptance}
\label{form-acc-sec} 

Dialogue \ex{1} is a financial advice dialogue about a way to reinvest the caller's
pension.  It illustrates the most minimal forms for explicit acceptance, prompts such as
{\it uh huh}, {\it I see}, {\it Sure}, {\it Right}, and {\it okay} as in \ex{1}-34 and 37
\cite{Schegloff82}.\footnote{These kinds of phrases have also been called {\sc
acknowledgments} and {\sc backchannels} \cite{WS88,WW90,Traum94,Carletta92}.}

\enumsentence
{(33) H: Well, the amount that you have, the excess amount, the twenty
eight hundred \\ (34)R: Okay \\ (35)R: the amount that  was not
your own contribution,  you rollover. \\ (36) R: You rollover. \\ (37) H: 
Right....
\label{you-roll-examp}
}

Dialogue \ex{1} is an excerpt from a financial advice dialogue about how to
reinvest the caller's certificates of deposit as they come due.  Partial
repetitions such as \ref{put-that-examp}-27 are commonly used to indicate
acceptance. See also the partial repetition in \ex{0}-36.

\enumsentence{
(26) H: That's right. as they come due, give me a call, about a week
in advance.  But the first one that's due the 25th, {\it let's put
that into a 2 and a half year certificate} \\ (27) E: {\bf Put that in a 2
and a half year}.  Would ... \\ (28) H: Sure. We should get over 15
percent on that
\label{put-that-examp}
}

 An example of
the use of paraphrase to indicate acceptance is found in \ex{1}-20:

\eenumsentence
{\item[] (18) H: I see. {\it Are there any other children beside your
wife?} \\ (19) D: {\it No} \\ (20) H: {\bf Your wife is an only child}.  \\
(21) D: Right. And uh he wants to give her some security ..........
\label{only-child-examp}}

The final type of `no new information' forms for indicating acceptance are
utterances that make inferences explicit that follow from what has just been
asserted or proposed \cite{Walker92a}.\footnote{Both paraphrases and inferences
of course can have other effects as well as indicating acceptance, such as
changing perspective or point of view. Following traditional semantic
assumptions, these types of changes are not considered to be additional
information \cite{Barwise88a}.} Consider dialogue \ex{1}, where H makes an
inference explicit in \ex{1}-(17). This inference follows from the context that
the tax year under discussion is 1981, from the inference rule of {\sc modus
tollens}, and from the context that results from adding \ex{1}-15 and \ex{1}-16
to the original context.

\eenumsentence
{\item[] (15) H: Oh no.  {\it IRA's were available as long as you are
not a participant in an existing pension.}\footnote{I R A stands for Individual Retirement Account which is
a way of putting money aside tax free until some time in the future,
such as when the holder retires.} \\ (16) J: Oh I see. \\ Well
{\it I did work I do work for a company that has a pension} \\ (17) H:
ahh.  {\bf Then you're not eligible for eighty one}.
\label{elig-examp} }

Given this variety in form, and the real-time nature of conversation, we next ask
what distributional features speaker A might use in identifying an utterance by
speaker B as an acceptance.

\subsection{Distributional Analysis of Acceptances}

One cue that speaker A may use is that explicit acceptances are logically
consistent with what A has asserted or proposed \cite{WS88,WW90}. As discussed in
section \ref{intro-sec}, logical consistency is necessary for acceptance,
although it is not sufficient.  In order to determine what other cues speaker A
might use to recognize B's utterance as an indicator of acceptance, I first
defined a superclass of repetitions, paraphrases and making inferences that I
could use in tagging a corpus of dialogues. This superclass is the class of {\sc
informationally redundant utterances} (IRUs) \cite{Walker93c}:

\begin{quote}
An utterance $u_i$ is INFORMATIONALLY REDUNDANT in a discourse
situation $\cal S$ if $u_i$ expresses a proposition $p_i$, and another
utterance $u_j$ that entails, presupposes or implicates $p_i$ has
already been said in $\cal S$
\end{quote}

It will be useful in what follows to have a term that we can use to refer to the
utterance $u_j$ that originally introduced the propositional content of the IRU
to the common ground. This is called the IRU's {\sc antecedent}.\footnote{IRUs in
fact often have multiple antecedents in the discourse, e.g. the utterance that
made an inference explicit in \ref{elig-examp}. In these cases, the most recent
utterance is the IRU's antecedent.}

Once the class of IRUs was defined, I collected 184 instances of IRUs from the
financial advice dialogue corpus \cite{PHW82}. In order to determine what cues
speaker A might use to recognize B's utterance as an acceptance, I coded each IRU
for a number of distributional factors that were hypothesized as potential cues.
First, each IRU was coded for the logical relationship of the IRU with its
antecedent and for the distance of the IRU from its antecedent. It was
hypothesized that these factors might affect the realization of the IRU. In
addition, since explicit acceptances provide no new information, a plausible
hypothesis was that acceptances would be prosodically marked as old information
\cite{Prince81,Brown83,Terken85,PH90}.  Both phrase final tone and phrasal
intonational contour were coded.

The first thing that the corpus analysis makes
apparent is that IRUs have other discourse functions besides indicating 
acceptance. For example,
a speaker may elaborate his/her own contribution with an IRU as H does in 
\ex{1}:

\enumsentence{But separate it. I don't want it all in one.} 

 An
examination of IRUs that indicate
acceptance however shows that they are said by  speaker B immediately after
speaker A finishes speaking. Note that in dialogue \ref{you-roll-examp}, the 
{\sc antecedent}
of the IRU in utterance 36 is utterance 35. In dialogue
\ref{put-that-examp}, the antecedent is utterance 26 and the IRU is
utterance 27.  In \ref{only-child-examp}, the antecedents are
utterances 18 and 19, and the IRU is utterance 20.  In dialogue
\ref{elig-examp}, the antecedents are utterances 15 and 16, and 
the IRU is utterance 17.

Thus, if we simply classify IRUs by speaker, and by location with
respect to their antecedents, we find that acceptance IRUs can be
largely described by two distributional factors:
\begin{enumerate}
\item Adjacent: the IRU sequentially follows its antecedent
utterance, ie. the IRU is U$_{n+1}$ and its antecedent is U$_{n}$.
\item Other: the speaker of the antecedent of the IRU is different
than the speaker of the IRU. 
\end{enumerate}

This Adjacent plus Other position is what was called the {\sc attitude locus}
above;  IRUs that occur in this position are termed Attitude IRUs.
Figure \ref{att-dist-fig} shows that the Adjacent plus Other parameters divide
the corpus roughly into two categories, with 93 IRUs classified as Attitude
IRUs according to these parameters.

\begin{figure}[htb]
\begin{center}
\begin{tabular}{|r||c|c|c|}
\hline & & &\\
& Repetitions & Paraphrases & Inferences  \\ \hline \hline &&&\\
Attitude IRUs& 54  & 15  & 24\\
Not Attitude IRUs & 6 & 43 & 32\\
\hline 
\end{tabular}
\end{center}
\caption{Distribution of all IRUs. Attitude IRUs indicate 
the Other speaker's attitude toward a speaker's assertion or
proposal, and are Adjacent
to the assertion or proposal.}
\label{att-dist-fig}
\end{figure}

\begin{figure}[htb]
\begin{center}
\begin{tabular}{|r||c|c|c|c|}
 \hline  &&& \\
Boundary Tone &  High &  Mid &  Low \\ \hline \hline &&&\\
Attitude IRUs  & 24 & 28 & 15  \\
Not Attitude IRUs  & 3 & 22 & 32 \\
\hline 
\end{tabular}
\caption{Distribution of Final Tones on Attitude IRUs (Adjacent
plus Other) vs Other IRUs}
\label{att-bt-fig}
\end{center}
\end{figure}

Most, but not all, Attitude IRUs are explicit acceptances. The exceptions are
those whose phrase final tone is a high rise (question contour).  See figure
\ref{att-bt-fig}. This subset with a high phrase final tone were called {\sc
echoes} by Cruttenden, who states that an echo `queries the whole or some part of
a previous utterance' \cite{Cruttenden86}.  

The remainder of the IRUs in the attitude locus, which are realized with final
mid and low tones, all indicate acceptance.  Figure \ref{att-bt-fig} shows that
falls to mid are significantly more likely to occur on acceptances than on IRUs
that have other functions, ($\chi^2 = 5.695, p < .02, df = 1$).  This finding is
consistent with the claim that final mids mark non-assertion, non-completion, and
hearer-old or predictable information \cite{Ladd80,Liberman75,McLemore91}, and
means that it is plausible that in about half the cases, A can recognize that B's
utterance, in the attitude locus, indicates acceptance simply from the final tone.

\begin{figure*}[htb]
\centerline{\psfig{figure=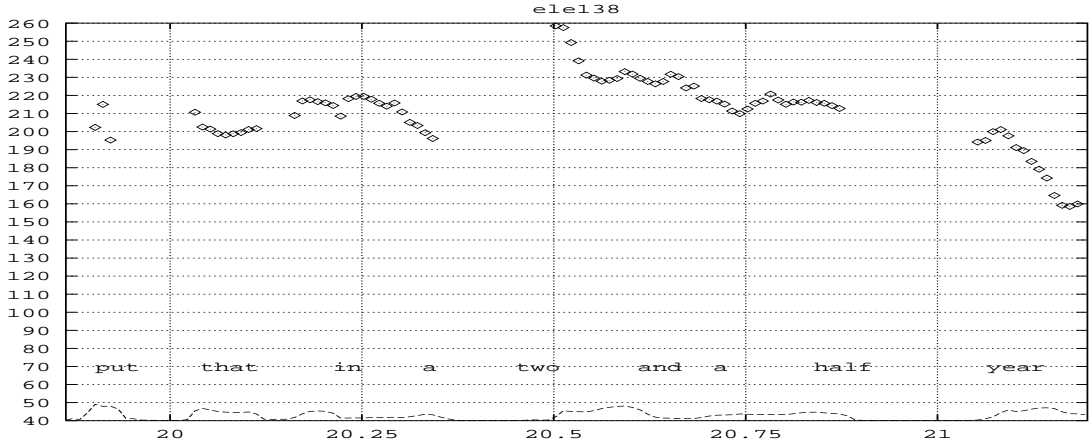,height=2.5in,width=6.5in}}
\caption{E's indication of acceptance by repetition from utterance 27 in 
dialogue \protect\ref{put-that-examp}. Y-axis is F0, X-axis is time.}
\label{ele138-fig}
\end{figure*} 

\begin{figure*}[htb]
\centerline{\psfig{figure=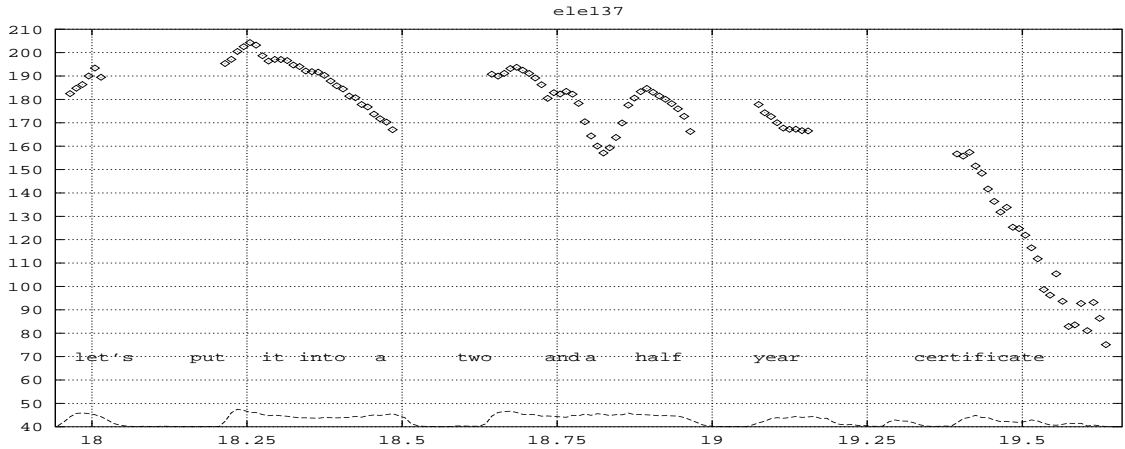,height=2.5in,width=6.5in}}
\caption{H's original assertion of utterance 26 in dialogue \protect\ref{put-that-examp}. Y-axis is F0, X-axis is time.}
\label{ele137-fig}
\end{figure*} 

Turning to phrasal intonational contour, the distributional analysis shows that,
in most cases, acceptances are treated as old information by being realized with
either sustained tones or with downstepped H*+L accents \cite{Pierrehumbert80,PH90}.
An example of a sustained tone is given in figure \ref{ele138-fig}, which is the
prosodic realization of utterance 27 in dialogue \ref{put-that-examp}.  The F0 of
this figure should be contrasted with that in figure \ref{ele137-fig} which shows
how the original assertion,  utterance 26 in dialogue \ref{put-that-examp},
was prosodically realized.  Note that {\it put it} was accented in Figure
\ref{ele137-fig} but that {\it put that} is not accented in Figure \ref{ele138-fig}.
Note also that the original assertion ends with a final fall to low whereas E's
acceptance IRU ends with a final fall to mid.

The repetition in 9 in dialogue \ex{1}, which indicates acceptance, illustrates
the use of downstepped H*+L accents:

\eenumsentence
{\item[](8) H: You can stop right there:  {\it take your money} \\
        (9) J: Take the money.   \\
        (10)H: Absolutely.....
\label{take-money-ex}
}

\begin{figure*}[htb]
\centerline{\psfig{figure=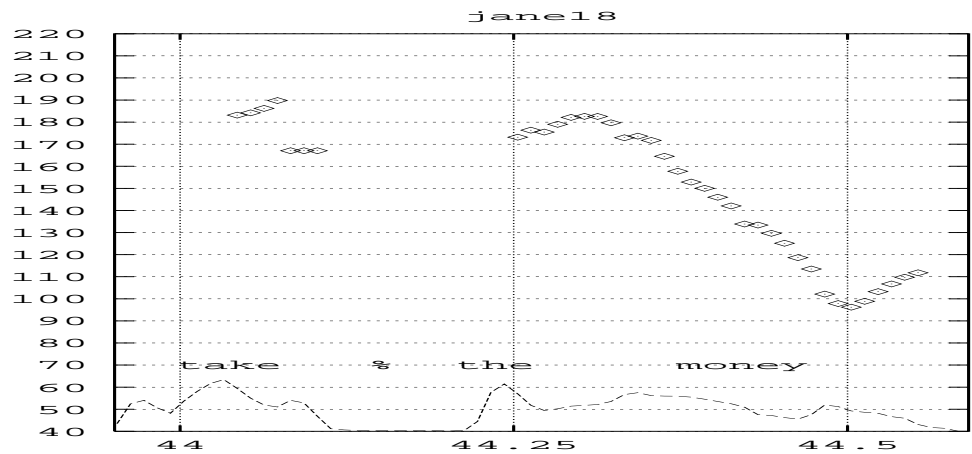,height=2.5in,width=6.5in}}
\caption{J's acceptance IRU in utterance 9 in dialogue 
\protect\ref{take-money-ex}. Y-axis is F0, X-axis is time. }
\label{jane18-fig}
\end{figure*}

\begin{figure*}[htb]
\centerline{\psfig{figure=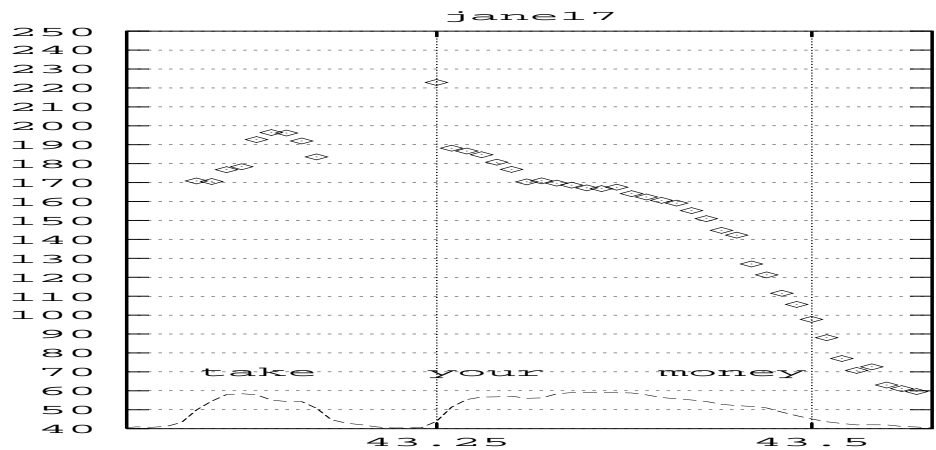,height=2.5in,width=6.5in}}
\caption{H's original assertion in utterance 8 in dialogue 
\protect\ref{take-money-ex}. Y-axis is F0, X-axis is time.}
\label{jane17-fig}
\end{figure*}

As shown in figure \ref{jane18-fig}, the utterance in \ex{0}-9 is
realized as a series of downstepped highs, with an H*+L pitch accent
first on {\it take} and then on {\it money}.  The final tone is a mid.
Contrast figure \ref{jane18-fig} with the prosodic realization of of
{\it take the money} in H's original assertion in figure
\ref{jane17-fig}. In figure \ref{jane17-fig}, the verb {\it take}
receives the primary accent and this accent is a simple H*, and the
utterance ends with a phrase final low rather than a mid.

Thus, differences in both phrasal intonational contour and phrase final tone
indicate that acceptance IRUs will in general sound qualitatively different than
their original assertions. The differences in prosodic realization mean that if
original assertions were realized with identical prosody to that of acceptance
IRUs, these assertions might be interpreted differently. Furthermore, if
acceptance IRUs were realized with the prosody of the original assertions, they
would probably not be interpreted as acceptances. One ramification of these
differences is that it is plausible that speaker A can identify these utterances
as acceptances without doing any logical processing at all; A may simply listen
to their prosodic realization.

Finally, consider figure \ref{att-bt-fig} again, and the 27 IRUs with high
boundary tones, called {\sc echoes} by Cruttenden.  The existence of
these echoes is strong evidence for a conventional assumption of an {\sc attitude
locus}, as discussed above. If we assume that Attitude IRUs with high boundary
tones (final rises) conventionally indicate non-acceptance, then Figure
\ref{att-bt-fig} shows that the vast majority of IRUs said with high boundary
tones are in the {\sc attitude locus} ($\chi^2 = 17.68, p < .001$).  Furthermore,
the three IRUs in figure \ref{att-bt-fig} with final highs that are {\bf not} in
the {\sc attitude locus}, are explicitly marked as being out of place, as in the
utterance of {\it Oh hang in there my friend} marking H's paraphrases with phrase
final rises in \ex{1}:

\eenumsentence{
\item[]
M:  ....
    First of all, you
    know, ah there's an outfit here in Philadelphia that you know,
    that I put money in at a certain interest, and I can I can borrow
    on it at one percent more. And {\it it's a 30 month deal}, so I have
    not received any interest on it. You know I don't have to show
    any interest on it {\it because they have not given me you know any
    1099's or anything}, but.. \\
H: Oh hang in there my friend \\
M: I'm hanging in there. \\
H: {\bf You have a 30 month certificate?}  \\
M: Right. \\
H: {\bf And they have not sent you a 1099}?
}

This supports the claim that A can use the fact that B's acceptance or rejection
is expected in the {\sc attitude locus} to make default inferences about
acceptance. In addition, both the utterance's position in the attitude locus, and
the prosodic realization function as primary cues for speaker A that speaker B's
utterance indicates acceptance.

\subsection{Summary of Acceptance}

This section first discussed conventions of dialogue such as the existence of the
attitude locus and the collaborative principle, that provide the basis for the dialogue
model assumed here. Then, I showed how utterances with `no new information' in the
attitude locus function as indicators of acceptance, and argued that these utterances
assert understanding while merely implicating acceptance. next, I used the distributional
analysis of acceptances to provide support for the existence of the attitude locus, and
to show that sequential organization and prosodic realization are primary cues that
speakers in dialogue can use to infer that an utterance implicates acceptance. In section
\ref{lewis-sec}, I show how the model of the common ground models the function of these
explicit acceptances in asserting understanding and implicating acceptance.

\section{Inferring Rejection in Dialogue}
\label{reject-sec}

In section \ref{intro-sec}, I introduced Horn's classification of rejection
\cite{Horn89}. In this section, I discuss how the naturally occurring examples of
rejection that I have collected from the literature and from the financial advice
corpus fit into Horn's classification.  It will be convenient to divide
rejections into two types: rejections of assertions and rejections of proposals.
In 55 dialogues, I found only 19 examples of rejecting assertions and 12 examples
of rejecting proposals.  While few in number, these examples highlight three
problems with previous work:

\begin{enumerate}
\item The literature mainly discusses the logical inferences required
to recognize the rejection of an assertion and
doesn't specify the basis for inference of the rejection of proposals (advice, offers etc.)
\item Previous classifications of  rejection are incomplete.
\item Types of rejections not previously noted pose a problem for the
assumptions and mechanisms of current theories about how to
represent the common ground.
\end{enumerate}

First, in section \ref{assert-rej}, I discuss the forms of rejecting assertions,
and then in section \ref{propose-rej}, I discuss the forms for rejecting
proposals.  In each example, the rejection form is indicated in {\bf boldface},
while the assertion or proposal that it rejects is indicated in {\it italics}.

\subsection{Rejecting Assertions}
\label{assert-rej}

The simplest cases of rejecting assertions are the types of denial and
contradiction discussed in previous work. In this small corpus, no examples of
implicit denial were found, but both contradictions and denials are
represented. Since these are straightforward, I will not discuss them
further. The other types of rejecting assertions are categorized into two
classes. The first class are those where the inference of rejection is based on
the content of the rejection itself, but the inference of rejection follows from
an implicature as in example \ref{garage-examp}. I call these {\sc implicature
rejections}. The second class relies on drawing inferences based on epistemic
inference rules; these are dubbed {\sc epistemic rejections}.

\subsubsection{Rejection by Implicature}
\label{implic-rej}

The class of {\sc implicature rejections} is illustrated by example
\ref{garage-examp}, where B's rejection shows that logical inconsistency is not
necessary for rejection. So how does A infer that B's utterance is a rejection?
A plausible, and I argue, the correct analysis, is that B's response rejects A's
by the process of scalar implicature, similar to the implicature shown in \ex{2},
generated by the fact that \ex{1}B is 'less informative' than it might have been
\cite{Hirschberg85,Gazdar79,Horn72}:

\eenumsentence
{\item[A:] Is the new student brilliant and imaginative?  \item[B:]
He's imaginative.
\label{brill-examp}
}

\enumsentence{
He's not brilliant.
\label{brill-implic}
}

In the remainder of this section, I present the relevant aspects of Hirschberg's
theory of scalar implicature \cite{Hirschberg85} in order to show how the
inference of rejection is derived in \ref{garage-examp}, and discuss the issues
that arise from this analysis.

\paragraph{Hirschberg's theory of Scalar Implicature}
Quantity implicatures arise from the use of a less informative item in
a sentence, implicating that the same sentence with a more informative
item is either false or unknown.  Thus the less informative {\it some}
in \ex{1} gives rise to the implicature in \ex{2}:

\enumsentence
{Kim ate some of the cookies.
\label{cookie-examp}}

\enumsentence
{Kim didn't eat all of the cookies.
\label{cookie-implic}}

Scalar implicatures are a type of quantity implicature \cite{Horn72,Gazdar79}.
Hirschberg's theory of scalar implicature specifies the conditions under which a
speaker may {\sc license} a scalar implicature and that a hearer must have access
to in order to {\sc infer} that a speaker intended a particular scalar
implicature \cite{Hirschberg85}.  For the class of {\sc implicature rejections},
we will see below how Hirschberg's conditions on licensing and inferring
implicatures apply directly to the licensing and inferring of rejection
implicatures.

Scalar implicatures are calculated from surface
semantic representations of propositions, i.e. from logical form, by
identifying a potentially scalar subformula in the logical form,
identifying the scale or scales that this subformula belongs to, and
generating implicatures for alternate and higher values of that scale.
The theory depends on: 

\begin{enumerate}
\item $\cal O$, a salient {\sc ordering} or scale,
defined as any partially ordered set, {\sc poset}, relation over a set
of scalar expressions $e_1 \ldots e_n$; 
\item  a means of ranking
sentences as {\sc higher}, {\sc lower} or {\sc alternate} sentences
with respect to $\cal O$; and 
\item  a specification of whether the
speaker uttered a sentence which {\sc affirmed}, {\sc denied} or
declared {\sc ignorance} of a value on $\cal O$.
\end{enumerate}

A means of ranking sentences is provided by the definition of a scale
as a {\sc poset}, e.g. a {\sc higher} sentence is a sentence with a
higher value from the {\sc poset}.  The expressions $e_i$ which can
participate in scales are any constant, predicate, logical or
epistemic operator, connective or quantifier symbol of a proposition
$p_i$ or any wff that is a subformula of $p_i$.  Here we only consider
sentences that {\sc affirm} a value $e_i$ on a scale $\cal O$ in an
asserted proposition $p_i$, as defined in \ex{1}.  A sentence $p_i$ is
{\sc simple} with respect to an occurrence of a component expression
$e_i$ iff $p_i$ contains no instances of negation with wider scope
than $e_i$.

\enumsentence{
AFFIRM(S, $e_i$, $p_i$) iff ($p_i$ = BEL(S, $p_j) \wedge$
SIMPLE($p_j,e_i$)) }

As \ex{1} shows, Hirschberg's formulation assumes that every utterance
can be represented as S's commitment to belief in some proposition or
to lack of such belief. In other words an utterance which realizes a
proposition $p_i$ is represented as BEL(S,$p_j$) or
$\neg$BEL(S,$p_j$).

The {\sc scalar implicature inference rule} (SIIR) for {\sc affirmed}
sentences is in \ex{1},\footnote{See \cite{Hirschberg85} for inference
rules for utterances which express denial and and declaring
ignorance.}  where $\cal O$ is an ordering, $C_h$ is the context, and
$BMB$ is the standard Belief in alternating mutual belief:

\enumsentence{
{\sc Scalar Implicature Inference Rule}(SIIR):\\ $\exists {\cal O}
(BMB( Salient({\cal O} ,C_h) \\ \wedge Realize(u_i, Affirm(S, e_i,
Bel(S,p_i)) \\
\wedge (HigherSent(p_i, p_j, {\cal O}) \vee AltSent(p_i, p_j, O))) \\
\rightarrow ScalarImp(S, H, u_i, \neg BEL(S, p_j), C_h)))$
}

The SIIR says that if there is a scale $\cal O$ that is salient in the
context and a speaker S affirms a sentence $p_i$ with a component
expression $e_i$, and there is another sentence $p_j$ which is a
higher sentence or alternate sentence to $p_i$ with respect to scale
$\cal O$, then the speaker may implicate that it's not the case that
s/he believes $p_j$, i.e. either s/he doesn't know whether $p_j$ or
she believes not $p_j$.\footnote{The rule is based on a three-valued
logic. The denial ($\neg$ T) of one of three logical
possibilities (T,F,\#) in a three valued logic is equivalent to the
disjunction of the other two (F $\vee$ \#). Note that $\neg$ Bel(S,
p$_j$) will be true in a two-valued logic just in case BEL(S,
$\neg$p$_j$) and similarly, $\neg$BEL(S,$\neg$p$_j$) will be true just
in case BEL(S,p$_j$). In other words, where logical systems do not
permit the representation of ignorance, scalar implicatures may still
be represented as simplified by the assumption of these systems in the
same way that ignorance is simplified by them \cite{Hirschberg85}, p.
81.}

Figure \ref{scales-fig} illustrates potential sample scales $\cal O$
\cite{Horn89,Hirschberg85}. Some of these can be defined by entailment
such as the conjunctive assertion scale: $P \wedge Q$ entails the
truth of $P$ and of $Q$. Others must be based on common knowledge of
the world: (VW, Opel, Honda, Chevy).  Still others are based on
common knowledge between the speaker and hearer and only constructed
for that particular interaction, e.g. (a dog, a stove) as a set of
{\it things we can afford} \cite{Ladd80}.

\begin{figure}[ht]
\begin{center}
\begin{tabular}{|l|}
\hline \\
SAMPLE SCALES $\cal O$ \\ \hline \hline \\
(definite, indefinite) \\
(all, most, many, some, few) \\
(necessarily, probably, possibly...)   \\
(.... ten, nine, eight....)   \\
(must, should, may, ...)  \\
(excellent, good)  \\
(hot, warm)  \\
(always, often, sometimes,...)  \\
(succeed in Ving, try to V, want to V)  \\
(love, like, don't mind)  \\
(none, not all) \\
($P, P \wedge Q$ ) \\
(apples,bananas,pears,plums,oranges) \\
(VW, Opel, Honda, Chevy) \\
(a dog, a stove)  \\
(a book, half of a book, a chapter of a book) \\
\hline 
\end{tabular}
\label{scales-fig}
\end{center}
\end{figure}

Thus if we instantiate the SIIR by letting $u_i$ be the assertion in
\ref{cookie-examp}, the scale $\cal O$ be the scale of quantifiers
({\it all, most, many, some, few}), the higher sentence $p_j$ be \ex{1}, and the
context $C_h$ the null context, we get the implicature in \ex{2},
glossed in \ref{cookie-implic}.

\enumsentence{Kim ate all of the cookies.}

\enumsentence{ScalarImp(S, H, {\it Kim ate some of the cookies},  \\
$\neg$ BEL(S, {\it Kim ate all of the cookies}), $C_h)))$ }

The predicate ScalarImp in \ex{0} is defined so that implicatures can
be felicitously {\sc cancelled}, as in \ex{1}a, as well as {\sc
reinforced} as in \ex{1}b.

\eenumsentence
{
\item Kim ate some of the cookies, and in fact Kim ate all of them.
\item Kim ate some of the cookies, but Kim didn't eat all of them.
}

The tests of {\sc cancellability} and {\sc reinforceability}
distinguish conversational implicatures from entailments
\cite{Grice75,Hirschberg85}.  In \ex{0}b, the implicature
was cancelled by a subsequent statement, but can also be cancelled by
{\bf prior} context, so that the implicature never arises, as in
\ex{1}.

\enumsentence
{Kim didn't eat all of the cookies. She ate some of them.
\label{prior-examp}
}

Thus in every respect, scalar implicatures have the logical properties of normal
{\sc default} inferences \cite{Perrault90}.  The remainder of the paper assumes
that the consequent of the SIIR has the logical status of a default, although
unlike defaults, scalar implicatures must be specifically {\sc licensed} by
features of the context \cite{Hirschberg85}.

\paragraph{Applying Hirschberg's theory to the examples}

Consider how this theory explains the other examples we have seen.  In
\ref{brill-examp}, A introduces a question as to whether $a$, {\it the
new student}, is both $brilliant$ and $imaginative$.  The conjunction
evokes the scale of conjunctive assertions ($P, P \wedge Q$ ), where
$P$ is {\it The new student is imaginative} and $Q$ is {\it The new
student is brilliant}, and $P \wedge Q$ is a higher sentence than $P$.
Thus because speaker B affirms $P$ with \ref{brill-examp}B, B
implicates the denial of $brilliant(a)$ in \ref{brill-implic}.

Now, most discussions of implicatures consider only question contexts such as
\ref{brill-examp}; out of the 185 naturally occurring examples in Hirschberg's
thesis, 180 implicature generating responses are responses to a question. But
note that the implicature shown in \ref{brill-implic} still arises in the context
of the assertion in \ex{1}A, showing that the implicature in \ref{brill-implic}
is not dependent on the question context given in \ref{brill-examp}.

\eenumsentence
{\item[A:] The new student is brilliant and imaginative.
\item[B:] He's imaginative.
\label{brill-examp-ass}
}

Thus, we can use the SIIR to license the inference of rejection in
\ref{garage-examp} if we take the salient scale  $\cal O$ to be ({\it a
man, something}). Then \ref{garage-examp}A is a higher sentence than
\ref{garage-examp}B and  B's assertion implicates that it isn't the
case that B believes \ref{garage-examp}A.  

Furthermore, these implicatures are understood as rejections
as long as a salient scale can be identified. In \ex{1}, from
\cite{Hirschberg85}, the salient scale is (love, like, don't mind).
Again note that \ex{1}B is logically consistent with \ex{1}A since {\it
like} entails {\it don't mind}.

\eenumsentence{
\item[A:] She likes it.
\item[B:] I don't mind it.         (Hirschberg's (125))
}

\paragraph{Issues with Implicature Rejections}

If scalar implicatures are generated in the context of an assertion,
then two issues arise. 

The first issue is that a less informative U$_2$ following an
assertion U$_1$ may {\sc accept} U$_1$ rather than {\sc reject} it, as
exemplified by example \ref{num-examp}. After discussing the remaining
types of rejection below, Section \ref{focus-sec}  discusses how
differences in information structure can be used by the hearer as cues
for distinguishing acceptance and implicature rejection.

The second issue is cancellability: implicatures only
arise when they are consistent with the context as shown by example
\ref{prior-examp}. Yet \ref{brill-implic} is not consistent with
\ref{brill-examp-ass}:A, so \ref{brill-examp-ass}:A cannot have been
added to the context as an assertion before the utterance of
\ref{brill-examp-ass}:B. I leave this issue aside until section
\ref{lewis-sec}, to be dealt with by the model of the common ground
introduced there.

\subsubsection{Epistemic Rejections}

The types of rejections discussed so far all rely on detecting logical
inconsistency or scalar implicatures by inference rules that operate
within the domain of the content of what is being discussed. There are
an additional three classes of rejecting assertions whose
identification relies on detecting conflicts that result from the
application of default epistemic inference rules.

Epistemic inference rules are rules of epistemic logic by which an agent can make
inferences about other agents' beliefs.  While there are many possible systems of
epistemic inference, here I draw on a set of default rules used by both Perrault
and by Appelt and Konolige in their work on expressing the effects of speech acts
as defaults of an epistemic inference system
\cite{Perrault90,AK88}. \footnote{For convenience I have short circuited the
inference system so that Belief Transfer is not mediated by a rule by which agent
B can infer from a declarative utterance by speaker A that speaker A believes the
content of that utterance. With that system the Belief Transfer Rule has speaker
A's belief as the antecedent, i.e.  speaker A's belief is the reason for speaker
B to adopt a belief in the proposition p, rather than speaker A's assertion as we
do here.} My assumption is that these default rules of epistemic inference have
the same logical status as implicatures, and that the standard problem of
conflicting defaults can arise between these inferences and implicature
inferences as specified by the SIIR. See also \cite{JWW86}. The first rule is
given in \ex{1}:

\enumsentence{ {\sc Belief Transfer Rule}: \\
Say(A,B,p) $\rightarrow$ Bel (B,p) }

The Belief Transfer Rule states that if one agent A makes an assertion
that p then by default another agent B will come to believe that p.
The second rule is in
\ex{1}: 

\enumsentence{ {\sc Belief Persistence Rule}: \\
Bel (B,p,t$_0$) $\rightarrow$ Bel (B,p,t$_1$) \label{bel-persist-rule}
}

The Belief Persistence Rule states that if an agent B believes p at
time t$_0$ that by default agent B still believes p at a later time
t$_1$.

These rules provide the basis for inferring three additional types of
rejections:

\begin{enumerate}
\item Denying Belief Transfer:  agent B can deny the consequent of the
Belief Transfer Rule by expressing doubt as to the truth of A's
assertion or by negatively evaluating A's assertion as to its
'sensibility'
\item Asserting  Inconsistent Past Belief: agent B can create a conflicting
default with Belief Transfer and Belief Persistence by asserting a
past belief which, if it persisted, would be inconsistent with the
results of the Belief Transfer Rule.
\item Cite Contradictory Authority: agent B can create conflicting defaults with
the Belief Transfer rule by stating another assertion attributed to
another agent, whose content contradicts what A has just asserted.
\end{enumerate}

Below I will give examples of each of these types.

\paragraph{Denying  Belief Transfer}

Conversant B can deny the consequent of the Belief Transfer Rule by
expressing doubt as to the truth of A's assertion 
as H does in \ex{1}
in utterance 14:

\enumsentence{
(11) B: Well ah {\it he uh ... he belongs to a money market fund now
and uh they will do that for him.} \\ (12) H: The money market fund
will invest it in government securities as part of their individual
retirement account -- is that what you're saying? \\ (13) B: Right. \\
(14) H: {\bf I'm not so sure of that.}
\label{mmfund-examp}
}

The hearer can also deny the effects of the Belief Transfer Rule by negatively
evaluating A's assertion as to its 'sensibility', as in utterance
12 in \ex{1}:

\enumsentence{
( 8) H: How much are you talking about? \\
( 9) E: About 65 thousand dollars. \\
(10) H: And if you are to take it periodically, what would that give
    you? \\
(11) E: {\it I don't know - nobody seems to be able to give me any kind of
    an answer.} \\
(12) H: {\bf That doesn't make sense}.
}

\paragraph{Inconsistent Past Belief}

Inferring that B's expression of an inconsistent past belief is a type
of rejection relies on detecting conflicting defaults with the Belief
Transfer Rule and the Belief Persistence Rule.  The simplest case is
shown in \ex{1} and \ex{2} where two beliefs directly conflict, but
the relevant past belief can be any inconsistent belief q where q
$\rightarrow$ $\neg$ p is a mutually believed rule of inference.

\enumsentence{
Say(A,B,p) $\rightarrow$ Bel (B,p)}

\enumsentence{
Bel (B,$\neg$ p,t$_0$) $\rightarrow$ Bel (B,$\neg$p,t$_1$)}

An expression of an inconsistent past belief is shown in dialogue
\ex{1} in M's utterance in 13:

\enumsentence{
( 6) H: You have a 30 month certificate?  \\ ( 7) M: Right . \\ ( 8)
H: And they have not sent you a 1099? \\ ( 9) M: No, .... \\
.....\\
(12) H: {\it Then they are remiss in not sending it to you} because
that money is taxable sir.\\
 (13) M: I know it's taxable, {\bf but I
thought they would wait until the end of the 30 months.} \\ (14) H: No
sir......  }

\paragraph{Citing Contradictory Authority}

Inferring that citing a contradictory authority is a type of rejection
relies on recognizing two inconsistent instantiations of the Belief
Transfer rule as shown in \ex{1} and \ex{2}:

\enumsentence{

Say(A1,B,p) $\rightarrow$ Bel (B,p)}

\enumsentence{ Say(A2,B,$\neg$p)  $\rightarrow$ Bel (B,$\neg$p)}

In other words, conversant A1 asserted p and conversant A2 asserted $\neg$p,
potentially leaving B in an inconsistent belief state caused by the
conflicting defaults generated by the alternate instantiations of the
Belief Transfer Rule.  As in the case of inconsistent past beliefs,
any conflicting belief q where q $\rightarrow$ $\neg$p is a mutually
believed rule of inference can generate a conflicting default of this
type. 

In the corpus of rejections, citing contradictory authority is used to
reject the talk show host's assertion three times, with the local
bank, the managers of a money market fund, and the IRS all serving as
alternate sources of expertise.  An example is found in \ex{1} where,
in a continuation of dialogue \ref{mmfund-examp}, B asserts in 15 that
the managers of the money market fund support his original assertion
in 11, thereby indicating that he rejects H's rejection in
14:\footnote{Notice that H treats B's assertion as a rejection by
repeating his statement of doubt in 16, and then elaborating on his
contradictory beliefs.}

\enumsentence{
(15) B: {\bf That's what they told me.} \\ (16) H:  I'm not so
sure of it. \\ They may move it ah into a into a government
securities fund, but I'm not so sure that they can move it into
individual securities -- check that out .  }

\subsection{Rejecting Proposals}
\label{propose-rej}

The rejection of proposals is analyzed separately from rejecting assertions
for two reasons: (1) rejecting proposals have received less attention
in the literature; and (2) inferring proposal rejection assumes
inferences based on processes of means-end reasoning and deliberation
rather than logical reasoning.

The first observation from the corpus analysis of rejecting proposals is that
there is a parallel between rejecting assertions and rejecting proposals in terms
of inferences based on content in the domain vs.  those based on propositional
attitudes. I discuss the content class first, and then discuss the second class,
which I call {\sc deliberation rejections}.

\subsubsection{Refusal, Negative Consequence and Precondition Denial}

The first observation about rejecting proposals based on content is
that the rejecting assertions types of denial, contradiction and
presupposition denial have correlates in the action domain. These are
respectively:

\begin{enumerate}
\item Refusal (with or without a reason for refusing)
\item Asserting the negative consequence of a course of action proposed
\item Precondition denial
\end{enumerate}

Refusal is similar to denial because it straightforwardly rejects the
offer, advice or proposal. An example is given in \ex{1}, where in
utterance 41, R refuses to accept the suggested course of action that H
proposes in 38, and includes the reason for the refusal.

\enumsentence{
(38) H: {\it And I'd like 15 thousand in a 2 and a half year
certificate }\\ (39) R: The full 15 in a 2 and a half?  \\ (40) H:
That's correct. \\ (41) R: {\bf Gee. Not at my age.  } }

The similarity of negative consequence to contradiction rests on the
assumption that means-end reasoning in the action domain is equivalent
to logical entailment in the truth-conditional domain. Thus asserting
the negative consequence of a course of action is equivalent to
asserting a logically inconsistent fact, a contradiction.  

In dialogue \ex{1}, the talk show host H, and the caller D, are discussing an
arrangement whereby D's father-in-law would buy a house for D and his wife. H's
advice, as summarized in 26 and 28 is for D to buy the house himself by borrowing
the money from his father-in-law.  D's utterance in 29 leads to an inference of a
negative consequence that would arise from following H's advice, the advised
course of action would not result in an investment for D's father-in-law.

\enumsentence{
(26) H: Well I don't like that kind of ownership.  \\ I don't like it
from your father-in-law's point of view \\ and I don't like it from
your point of view. \\ If your father-in-law chooses and if you decide
to do it, {\it let him lend you the money, }\\ (27) D: mhm \\ (28) H:
and then you can purchase the place,  keep the place up,  pay him
a fair rate of return on his money,  and pay him back \\ (29) D:
{\bf Well  he kind of wants it as sort of an investment.} \\
(30) H: I'd rather see him invest elsewhere.  \\ }

Precondition denial results from a situation in which advice is given that
presupposes that a precondition for doing the advised action or for accepting the
advice holds.  A rejection of this type is simply a statement that a precondition
for the advised action does not hold, or a statement from which it is inferrable
that a precondition for the advised action does not hold.  In \ex{1}, H's
statement in 13 is an indirect form of a proposal that J should have an I R A for
last year. The rejection in 14 is of the second type: the denial of the
precondition must be inferred via the rule of Belief Persistence given in
\ref{bel-persist-rule} from what is actually stated in 14:

\enumsentence{
(13) H: {\it And there's no reason why you shouldn't have an I R A for last
year.}\\ (14) J: {\bf Well but I thought they just
started this year.}  \\ 
(15) H: Oh no. IRA's were available as long as
you are not a participant in an existing pension.  }

\begin{figure}[htb]
\centerline{\psfig{figure=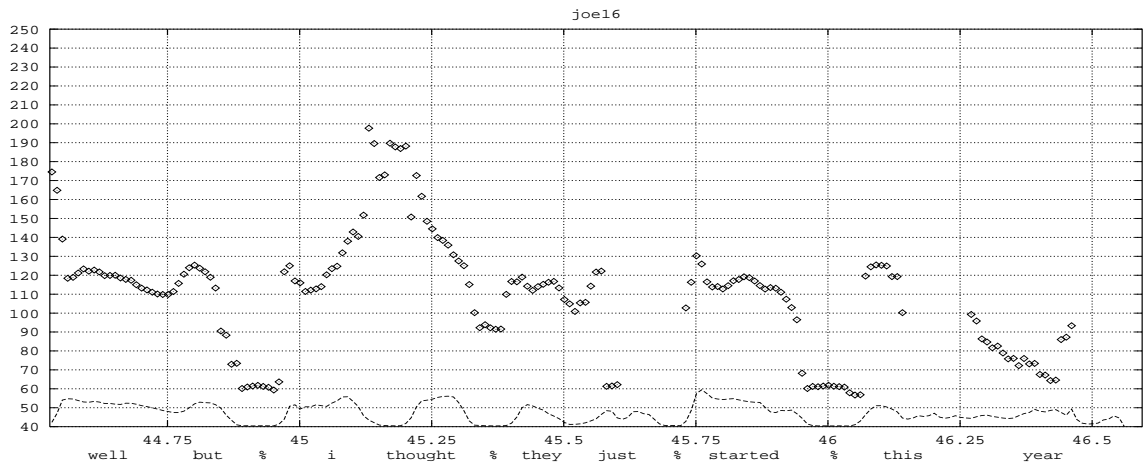,height=2.5in,width=6.5in}}
\caption{Joe 14. Rejection, Focus Marked. Y-axis is F0, X-axis is time.}
\label{thought-rej-fig}.
\end{figure}

H's process of inferring that J's utterance counts as a rejection may be assisted
from the prosody with which J realizes his utterance.  See figure
\ref{thought-rej-fig}, where the accent on {\it thought} marks its complement as
factual \cite{Cruttenden86}.

\subsubsection{Deliberation Rejections}

Deliberation rejections are a way of rejecting proposals that rely on default
inferences about the deliberative processes by which an agent evaluates the
potential benefit or utility of pursuing a course of action \cite{BIP88,Doyle92}.
Deliberation inferences are the action parallel of epistemic inferences.  Where
epistemic inferences are concerned with belief and truth, deliberative inferences
involve intentions and the desirability, utility or expected payoff of one
potential intention when compared with another. Rejections based on deliberation
rely on default rules of intention adoption and persistence such as those below:

\enumsentence{ {\sc Intention Adoption Rule}: \\ 
Propose(A,B,action$_i$) $\rightarrow$ Intend(B, action$_i$, degree$_i$) }

\enumsentence{ {\sc Intention Persistence Rule}: \\
Intend(B, action, t$_0$) $\rightarrow$ Intend(B,action, t$_1$) }

These rules reflect the assumption that conversants in situations such as the
financial advice situation commonly have some shared goal of finding a solution
to the caller's problem \cite{GS90,WW90,Traum94,ChuCarberry94}. Given this shared
goal, they then deliberate about which intentions to adopt as means for
satisfying that goal \cite{BIP88,Doyle92}.  The degree$_i$ in the consequent of
\ex{-1} is defined by possible courses of action, or alternate potential plans
that a conversant may consider, and the degree of the utility of the proposed
course of action in comparison with other alternate means.  These inference rules
are default inference rules (cf. \cite{JWW86,Reiter80,LAO92}), similar to those
for epistemic inference. They support two ways of rejecting proposals:

\begin{enumerate}
\item Negative Evaluation: Agent B negatively evaluates  the proposed course
of action, stating that it is not desirable.
\item Conflicting Intentions: Agent B proposes and possibly evaluates the 
desirability of another option generated as a means of satisfying the
same goal by means-end reasoning
\end{enumerate}

I give examples of each of these types below.

\paragraph{Negative Evaluation}

Negative Evaluation consists in denying the effect of the Intention Adoption Rule
by stating that the degree of expected utility is low.  Dialogue \ex{1},
utterance 79 is an example of Negative Evaluation, given in response to the
proposed course of action in 78:

\enumsentence{
(78) H: {\it Put this money aside in treasury notes. }\\ (79) M: Now
wait, I want..uh {\bf in a 43 tax bracket I didn't think that was too
wise, } .....\\ }

\paragraph{Conflicting Intentions}

Conflicting intentions arise when conversant B proposes and possibly evaluates the
desirability of another option generated as a means of satisfying the same goal
by means-end reasoning.

Dialogue \ref{consider-cash-examp} illustrates conflicting intentions inferred by
the Intention Persistence Rule.  In dialogue \ex{1}, H and M are discussing
getting some money from a life insurance policy and potentially increasing the
level of coverage afterward.  Intention persistence is exemplified by M's
assertions in 55: M uses the past tense in making her assertion about her
potential intentions, and H must infer that those intentions have persisted in
order to infer that M is rejecting his proposal to cash the policies in.

\enumsentence{
(52) H: So what I would suggest is that he apply for insurance to make
up that difference.  If indeed you feel you need the 12 thousand
dollars in insurance.  Do you need it?  \\ (53) M: Not really. \\ (54)
H: Well if you don't need it, {\it then let's cash the policy in} \\
(55) M: {\bf I wasn't considering cashing the policies, I was
considering borrowing the money.} \\ (56) H: Well let's play it both
ways. \\ (57) M: Ok. \\
\label{consider-cash-examp}}

\subsection{Rejection Summary}

This section discussed the inference of rejection in dialogue. I first
distinguished the rejection of assertions from the rejection of proposals. Then I
showed that both of these classes of rejections rely on default inference
rules. Rejecting assertions introduces defaults of two types: {\sc implicature
rejections} and {\sc epistemic rejections}. I showed how {\sc implicature
rejections} rely on inferences of scalar implicature based on the content of the
utterance, while {\sc epistemic rejections} rely on inferences from epistemic
inference rules. Then I argued that inferences about rejections of proposals rely
on the same distinction between inferences based on the content of the proposal,
and inferences based on meta-level rules of inference, deliberation inference
rules. I introduced and exemplified new classes of proposal rejection, based on
means-end reasoning and deliberation, and showed that these inference rules are
default rules of inference whose effects must be accounted for in a model of the
common ground.

\section{The Role of Information Structure in distinguishing Acceptance and Implicature Rejection}
\label{focus-sec}

An issue raised in section \ref{reject-sec} with the analysis of {\sc implicature
rejections} is that a less informative U$_2$ following an assertion U$_1$ may
{\sc accept} U$_1$ rather than {\sc reject} it, as exemplified by example
\ref{num-examp}. The treatment of this issue below relies on detecting regularities in
information structure as indicated by prosodic contour. Before presenting the
analysis, this section first briefly reviews theoretical background on
information structure.


It is well known that information in a discourse does not consist of an
unstructured set of propositions, but that speakers form their utterances to {\bf
structure} the information they wish to convey.  The basis for this {\sc
information structure} is the speaker's beliefs about what the hearer knows and
what is currently salient for the hearer \cite{Prince81,Prince86,WH85,Delin89}.
For example, the use of a sentential subject has been shown to correlate with the
speaker's belief that the proposition conveyed is salient shared knowledge, and
is what has been called a {\sc presupposition} \cite{Horn86,Prince81b}.

A speaker can indicate information structure through two means: intonational
contour or syntactic form. Furthermore, these two means often reinforce one
another, so that syntactic constructions which mark information structure are
also realized with particular intonational contours \cite{WS79}. 

One important type of information structure marking is the marking of
variable-containing propositions, which have been called {\sc open propositions}
\cite{Prince81b,Prince86},  or {\sc p-skeletons} \cite{Rooth85}. Variables in an
open proposition can either be marked by syntactic forms with a gap, such as
topicalizations or questions \cite{Prince86}, or by intonational contour, in
which the location of the variable is marked as {\sc focal} by prosodic accents
of various kinds \cite{WH85,WS79,Jackendoff72}.

It is also well known that focus marking can be ambiguous as to the extent of the
domain of the variable, but that possible domains can be predicted by using
Liberman and Prince's relational theory of stress
\cite{LP77,Hirschberg85}. Liberman and Prince's relational theory analyzes stress
as binary feature with values of {\sc weak} and {\sc strong}. If lexical items in
an utterance have {\sc weak} labels, and are prosodically marked as focal, then
the domain of the variable is restricted to that lexical item. This is called
narrow focus. If lexical items in an utterance have {\sc strong} labels, and are
prosodically marked as focal, then the domain of the variable includes all the
nodes of the syntactic tree for the utterance that include that lexical item. The
analysis below assumes as theoretical background the use of focus to indicate
information structure, and Liberman and Prince's theory for determining the
domain of the variable in an open proposition \cite{LP77,Ladd80}.

\subsection{Distinguishing Acceptance from Rejection}
As  noted above, an issue with the analysis of {\sc implicature rejections} is that
a less informative U$_2$ following an assertion U$_1$ may {\sc accept} U$_1$
rather than {\sc reject} it, as exemplified by example \ref{num-examp}.  It is
plausible that the logical form for \ref{num-examp} includes conjunction as shown
in \ex{1}:

\enumsentence{
(house x) ((belong-to Sue x) $\wedge$ (located x Chestnut St.))  }

Thus, by the same reasoning we used above in the analysis of {\sc implicature
rejections}, we could argue that B's utterance evokes the scale of conjunctive
assertions, which in this case includes the conjunction shown in \ex{0}, and
furthermore that B's utterance realizes a lower item on the scale, namely {\it
on Chestnut St.}, thus implicating that it is not the case that B believes
\ref{num-examp}A. But B's utterance does {\bf not} implicate that for all B knows
it is not {\bf Sue's} house that is on Chestnut St.

Furthermore, none of the naturally occurring types of explicit acceptance
mentioned in section \ref{intro-sec} generate implicature rejections, even though
 they are typically logically consistent utterances that realize a subpart of
the propositional content of the previous assertion.  What is the difference
between \ref{num-examp} and \ref{garage-examp}?

A possible explanation is that A's utterance in \ref{num-examp} has no explicit
conjunction, and that explicit conjunction is required to introduce the scale of
conjunctive assertions.  However, consider the naturally occurring example in
\ex{1}.

\eenumsentence
{\item [A:] We bought these pajamas in New Orleans for me.  \item [B:]
We bought these pajamas in New Orleans.
\label{new-orleans-examp}
}

Here, the implicature conveyed by \ex{0}B, glossed in \ex{1}a, and given in
\ex{1}b, can be explained most naturally by postulating a conjunctive
representation at the propositional level as in \ex{2}.

\eenumsentence
{\item But not for you.  
\item ScalarImp(B, A, {\it We bought these
pajamas in New Orleans}, $\neg$ BEL(B, {\it We bought these pajamas in
New Orleans for you.}), $C_h)))$ }

\enumsentence{
(pajamas x) ((bought e x) $\wedge$ (located e New Orleans) $\wedge$
(agent e WE) $\wedge$ (benefactor e ME)) }

Since the implicature in \ex{-1}B, which indicates rejection, is
generated {\bf without} explicit conjunction, this explanation
does not seem plausible.

However, previous work on the form of acceptance, discussed in more detail in
\cite{Walker93c,Walker93a}, provides a partial answer to this problem.  In most
cases of repetitions that indicate acceptance, the repeated subformula of the
propositional representation of U$_2$ was either (1) previously questioned, i.e.
syntactically marked as focal \cite{Prince86}, or (2) prosodically marked as
focal in U$_1$.  In dialogue \ref{put-that-examp}, the IRU in 27 repeats
information marked as focal in utterance 26, as shown in figure \ref{ele137-fig}.
The repeated {\it on Chestnut St.} of \ref{num-examp}B would be {\bf focal}
information in \ref{num-examp}A.  Furthermore, as discussed above in relation to
explicit acceptances, this repeated focal information is prosodically marked as
hearer-old information by the speaker's choice of both prosodic contour and
phrase final tones \cite{Prince92}.

Thus one key difference between acceptances and implicature rejections is their
information structure. Acceptances re-realize focal information from U$_1$ and
mark it as old information.  Rejections re-realize the open proposition from
U$_1$, and replace the focal item with a scalarly related item.  This suggests
that the basis for inference of rejection includes the condition in \ex{1}:

\enumsentence{{\sc substitution of focus condition}: 

If an utterance U$_2$ by a speaker B asserts (proposes) an alternate
instantiation of the salient open proposition contributed to the context by an
assertion (proposal) U$_1$ as uttered by a speaker A, and U$_2$ omits, or
provides an alternate or more general instantiation of the focused element $e_i$
of U$_1$, then U$_2$ {\sc rejects} U$_1$.}

It follows from the analysis of rejection by implicature that the focused
elements $e_i$ are precisely those that potentially license scalar
implicatures. Thus, conversants in dialogue can use the {\sc substitution of
focus condition} as the basis for determining that a scalar implicature has been
licensed by a speaker's response,\footnote{This condition appears to define B's
utterance as, what Hirschberg calls, a {\sc simple expression alternate}, with
the addition of the focus constraints. In earlier work, I called this the
exclusion of focus condition, because of examples like
\ref{new-orleans-examp}B. However, \ref{new-orleans-examp}B can be viewed as a
substitution of {\it in New Orleans} for {\it in New Orleans for me}, and this
view is more consistent with the use of the conjunctive scale for generating the
rejection implicature.} and then infer what has been rejected on the basis of the
implicature.

\begin{figure}[tb]
\centerline{\psfig{figure=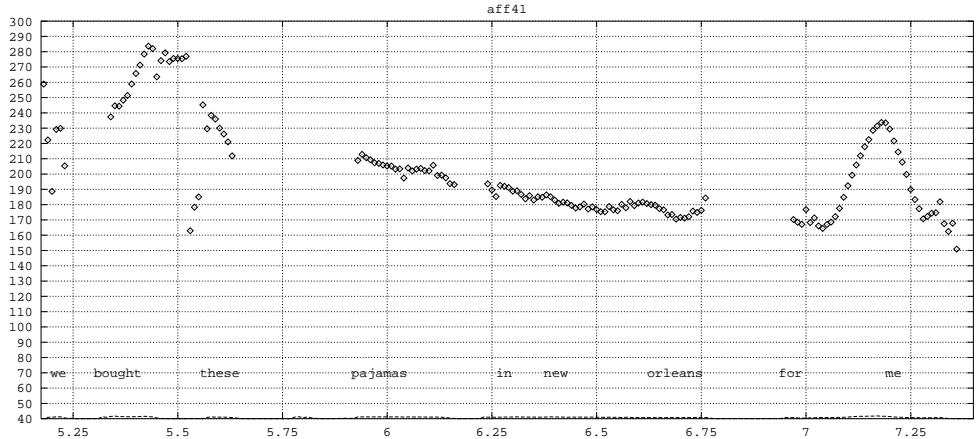,height=2.5in,width=5.5in}}
\caption{A natural focus marking on the  simple declarative in \protect\ref{new-orleans-examp}A. Y-axis is F0, X-axis is time.}
\label{aff41-fig}
\end{figure} 

Applying the {{\sc substitution of focus condition} to the examples thus far, we
see that in \ref{new-orleans-examp}A the focus is {\it for me}. See Figure
\ref{aff41-fig} as an example of the most natural way to mark focus in this
utterance.  In \ref{new-orleans-examp}B, B's replacement of this focus with one
that does not include {\it for me} leads to the construction of the conjunctive
scale, which minimally contains ((located e New Orleans) $\wedge$ (benefactor e
ME), (located e New Orleans)).  The conjunctive scale provides the basis for the
rejection implicature.

In \ref{garage-examp}A the focus is {\it a man} whereas in
\ref{garage-examp}B, the focus is {\it something}. The scale
of {\it a man, something} is made salient by the focus marking, and
{\it something} is a more general instantiation of {\it a man}, which
then licenses the rejection implicature.

In \ref{brill-examp}A and \ref{brill-examp-ass}A the focus is the
conjunction of {\it brilliant and imaginative}, whereas in
\ref{brill-examp}B and \ref{brill-examp-ass}B,  the focus is 
only {\it imaginative}. The rejection implicature is licensed by
identifying the scale of conjunctive assertions.

In all these cases B's assertion {\sc rejects} A's assertion because
it meets the {\sc substitution of focus condition}.

In contrast, \ref{num-examp}B realizes the focal element $e_i$ of
\ref{num-examp}A, failing to meet the {\sc substitution of focus
condition}. Thus \ref{num-examp}B accepts \ref{num-examp}A, as we
might have expected from the fact that it is logically consistent and
realizes $e_i$ as hearer-old information \cite{Prince92}.
\begin{figure*}[htb]
\centerline{\psfig{figure=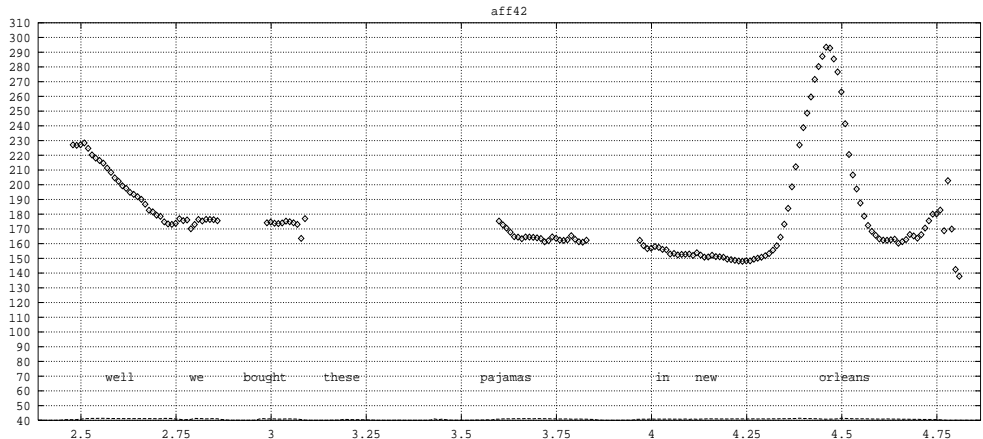,height=2.5in,width=5.5in}}
\caption{Fall Rise on B's rejection in dialogue \protect\ref{new-orleans-examp}. Y-axis is F0, X-axis is time.}
\label{no-pros-fig}
\end{figure*}

Furthermore, the {\sc substitution of focus condition} provides one explanation
of {\bf how} the speaker and hearer coordinate on which scales are salient, and
thereby coordinate their mutual beliefs about which scalar implicatures have been
licensed \cite{Moser92}.  The utterance expressions from which a scale is to be
identified are both marked as focal in U$_1$ and U$_2$. \footnote{This is similar
to, but different than a claim made in \cite{Rubinoff87} that only new
information licenses scalar implicatures, since old information may be marked as
focal.} Thus the focal structure of {\bf both} U$_1$ and U$_2$ is critical, for
defining the relevant scale $\cal O$, and for constraining when implicature
rejections are felicitous.  This can be seen by considering the difference in
foci of the naturally occurring \ref{new-orleans-examp}A and an invented
variation given here as \ref{new-orleans-examp}'A:

\eenumsentence[\ref{new-orleans-examp}']
{\item [A:] We bought me these pajamas in New Orleans.
\label{no-reject-examp}}

In the original utterance, the marked focus was {\it for me}. In
\ref{new-orleans-examp}' the focus is most natural on {\it New
Orleans}. B can reject \ref{new-orleans-examp}A with
\ref{new-orleans-examp}B by substituting an alternate focus, but B's utterance is
infelicitous as a rejection of \ref{new-orleans-examp}'A.  This is
precisely because the benefactor {\it for me} is not focal in
\ref{new-orleans-examp}'A.  It is not plausible that the propositional
representations from which scalar implicatures are calculated are
different in \ref{new-orleans-examp}A and \ref{new-orleans-examp}'A.
Thus the key factor is not simply that the propositional
representation makes available a scale of conjunctive assertions, the
scale must be identified from the focus/open proposition structure of
U$_1$ and U$_2$.

\subsection{Fall-Rise as an Indicator of Rejection}

An additional cue to distinguishing acceptances and implicature
rejections may come from the fact that implicature rejections are
often realized with a Fall-Rise intonational contour. Fall-Rise is a
contour which marks a focus, and which has been claimed to convey an
additional meaning as well. Figure \ref{no-pros-fig} shows the use of
Fall-Rise, as it would naturally be produced in utterance
\ref{new-orleans-examp}B.\footnote{Fall-Rise is a type of
falling-rising intonational contour describable in Pierrehumbert's
system as L$^*$+H$^-$L$^-$H\%. It is distinguished from other contours
by the fact that it is scooped \cite{Ladd80}.  There may be more than
one accented syllable in Fall-Rise, and for each such accented
syllable there must be an abrupt drop in pitch within the following
two syllables. In addition, Fall-Rise is characterized by a
sentence-final rise in pitch, as can be seen in figure
\ref{no-pros-fig}. This description is taken from \cite{Hirschberg85}
p. 49, see also \cite{WH85}.}

The intonational meaning of Fall-Rise, in addition to marking a focus,
has been characterized as incomplete deliberation \cite{Pike45} ,
uncertainty\cite{WH85}, focus within a set \cite{Ladd80}, a statement
or answer with reservation \cite{Halliday67}, a reminder that some
part of an utterance is background \cite{Gussenhoven84}, or a polite
softener of denial or rejection, \cite{Horn89,BrownLevinson87}.

In her thesis Hirschberg notes that some of her examples are most felicitous when
realized with Fall-Rise as Figure \ref{no-pros-fig} shows for utterance
\ref{new-orleans-examp}B.  Hirschberg proposed the use of Fall-Rise as a way of
indicating the salient scale in implicature, but stated that there were other
ways of indicating scales as well. In discussing example \ex{1}, Hirschberg
stated that such responses (rejections) are most felicitous with a pitch accent
or with Fall-Rise intonation over the indefinite in \ex{1}B:

\eenumsentence{
\item[A:] Works on the show.
\item[B:] Some show. (Hirschberg's (121) p 96)
}

Example \ex{0} is one of the 5 examples in Hirschberg's thesis of an
implicature generated in the context of an assertion; this suggests
that Fall-Rise is more likely to occur as a way to indicate the scale
in an implicature rejection than in other implicatures.

Ladd proposed that a Fall-Rise contour on the focused element signals
a subset or hyponym relation: the focused element represents a proper
subset or member of a contextually accessible set
\cite{Ladd80,Horn89}. In addition, he proposes that Fall-Rise signals
disagreement or at least amendment \cite{Ladd80}, p 155, i.e. \ex{1}B
disagrees with \ex{1}A, by marking {\it fool} with Fall-Rise, as
indicated in \ex{1}B. In contrast, the simple fall in \ex{2}B signals
acceptance by a continuation, i.e. conveys implicit acceptance.

\eenumsentence{
\item[A:]  Harry's the biggest liar in town.
\item[B:] The biggest $_{F}$fool$_{R}$, maybe (.. but I think he means
what he says) }

\eenumsentence{
\item[A:]  Harry's the biggest liar in town.
\item[B:] The biggest fool$_F$, maybe. (not only a liar but a fool)
\label{accept-liar-examp}
}

However, Fall-Rise is not {\bf necessary} because an utterance that
meets the {\sc substitution of focus condition} can reject without
Fall-Rise as in \ex{1}, even without the explicit amendation marker of
{\it maybe}.

\eenumsentence
{\item[A:] Sue's house is on Chestnut St.  \item[B:] Some street$_F$.
\label{num-examp3}
}

Furthermore, the Fall-Rise on \ex{1}B, which fails to meet the {\sc
substitution of focus condition}, is at best interpreted as a bit odd.
So Fall-Rise is neither necessary nor sufficient for indicating
rejection.

\eenumsentence
{\item[A:] Sue's house is on Chestnut St.  \item[B:] It's on
$_{F}$Chestnut$_{R}$ St.
\label{num-examp4}
}

However, in all likelihood, Fall-Rise plays an important role in helping A
disambiguate between rejection and continuation. It seems plausible that \ex{-1}B
could be ambiguous between continuation and rejection without the Fall-Rise or a
{\it maybe}.  If Fall-Rise indicates polite softening in this context
\cite{Horn89}, the presupposition of a less preferred response
(cf. \cite{BrownLevinson87}) would disambiguate between rejection and
continuation.

\section{Adding to the Common Ground}
\label{lewis-sec}

In sections \ref{acc-sec} and \ref{reject-sec}, I discussed the range of
ways that acceptance and rejection are indicated in dialogue. Modeling
the function of the various forms introduces several requirements on a
theory of how agents A and B remain coordinated on what is in the
common ground. These requirements are:

\begin{enumerate}
\item modelling the difference between implicit and explicit acceptance
\item explaining how implicatures can be used for indicating rejection
\item explaining how conflicting defaults are resolved in the inference of 
rejection
\end{enumerate}

In the remainder of this section, I present such a model.

\paragraph{Lewis's Shared Environment model of Common Knowledge}

The model proposed here is a computational version of Lewis's Shared
Environment model of common knowledge \cite{Lewis69,CM81}.
Previous work has noted that the common ground is more accurately
described as an individual speaker's `tacit assumptions'
\cite{Prince78b} or as `mutual absence of doubt'
\cite{Joshi82,NJ83}.  The {\sc mutual supposition} account of mutual
belief presented here models this `absence of doubt' quality by
representing the conversants' assumptions about mutuality as
defeasible, depending on the evidence provided by the other
conversants in dialogue.

Mutual suppositions can be modeled by an explicit schema proposed by
Lewis, called the {\sc shared environment} model of common knowledge
\cite{Lewis69,CM81}.  Because conversants don't have
access to the mental states of other conversants, mutual supposition
must be inferred, based on externalized behavior of various kinds.
The inference of mutual supposition can be inferred using the {\sc
mutual supposition induction schema}, henceforth MSIS:

\begin{quote}
{\bf Shared Environment Mutual Supposition Induction Schema (MSIS)} \\
It is mutually supposed in a population 
P that $\Psi$ 
if and only if some situation $\cal S$ holds such that:
\begin{enumerate}
\item Everyone in P has reason to believe that $\cal S$ holds.  
\item$\cal S$ indicates to everyone in P that everyone in P has reason to 
       believe that $\cal S$  holds.  
\item$\cal S$ indicates to everyone in P that $\Psi$.
\end{enumerate}
\end{quote}

In this work, the situation $\cal S$ of the MSIS is the discourse situation, and
the population P are the conversational participants.  If each of the three
conditions given in the MSIS is satisfied then the conversants are justified in
inferring that a fact $\Psi$ is mutually supposed.  Condition (1) specifies that
a public utterance event must be accessible to all of the discourse
participants. Conditions (2) and (3) state that what is mutually supposed is
derivable from the fact that all participants have access to this public
event. In other words, what is believed to be in the common ground is the set of
mutual suppositions {\bf indicated} by the occurrence of a sequence of utterance
events in a discourse situation $\cal S$.

Thus, according to the {\sc shared environment} model an utterance
event U in a discourse situation $\cal S$ licenses the inference of
certain mutual suppositions, depending on what U {\sc indicates} in
$\cal S$.  However what a discourse situation {\sc indicates} can vary
according to assumptions about language conventions and reasoning
processes.  

\paragraph{Making Lewis's Model Computational}

To formalize the {\sc indicates} relation, it is useful to augment the
representation of utterance events with two additional constructs:
{\sc assumptions} and {\sc endorsements}.
 
Assumptions are beliefs that support a {\sc mutual supposition}.  Let $\Delta$ be
a function on utterances that represents the set of {\sc defeasible assumptions}
associated with each utterance event.  In previous work, drawing from proposals
made by \cite{CS89,WS88}, and from Galliers' theory of belief revision
\cite{Galliers90,Galliers91a}, I suggested that one function of the various forms
of acceptance is to strengthen the evidence underlying the assumptions $\Delta$
associated with an utterance U \cite{Walker92a}.  $\Delta$ consists of
assumptions that speaker A makes that reflect A's expectations about the effects of
producing an utterance U:

\begin{enumerate}
\item {\sc attention
}: the
addressee B  attends to U; 

\item {\sc complete hearing}: the addressee hears U correctly; 

\item {\sc realize}: the addressee believes that U,
said in discourse situation $\cal S$, realizes a proposition $\Phi$,
and that the speaker intended to convey $\Phi$ .\footnote{Of course
the addressee may believe that U in $\cal S$ realizes some other
proposition besides the one that the speaker intended to convey.  The
realization assumption represents the fact that $\Phi$ is not conveyed
directly  \cite{Reddy79,Schegloff90,Brennan90}.  }

\item {\sc license}: the addressee believes that U,
 said in discourse situation $\cal S$, realizes
a proposition $\Phi$ and that the speaker intended the addressee
to infer $\Psi$, which follows from $\Phi$  as an entailment or
by non-logical inference. 
\item {\sc accept}: the addressee B believes that U
realizes a proposal or an assertion, and B accepts the
proposal or assertion, i.e. B intends the action proposed
or believes what has been asserted.
\end{enumerate}

In other words, in producing an utterance, a speaker assumes that the hearer is
attending, and that the hearer will understand the utterance, draw the intended
inferences, and accept what is asserted or proposed. These assumptions of the
speaker are very weak in terms of mutual beliefs, but can be reinforced by the
hearer's response, as we will see below.

The {\sc realize} and {\sc license} assumptions reflect the speaker's assumption
about the interpretation
process of conversants as they attempt to determine what the utterance U
indicates in a discourse.  For convenience, two functions are associated with
these assumptions. The function $\cal R$: ($\cal S$,U) $\rightarrow$ $\cal P$,
returns the proposition that the speaker of an utterance U in discourse situation
$\cal S$ intends to realize, and which the addressee must identify. The realize
function $\cal R$ is used for both assertions and proposals. The function $\cal
L$:~($\cal S, \rm{U}) \rightarrow \cal P$, returns the proposition licensed as an
inference by an utterance U in a discourse situation $\cal S$, which again the
addressee must identify.

We can now represent the inference of acceptance which leads to mutual
suppositions of beliefs and intentions as a default inference rule that depends
on the assumptions above. The {\sc indicates} function be represented as
$\leadsto_\Delta$ where $\Delta$ gives the set of associated assumptions, in the
Acceptance Inference Rule below, henceforth AIR.  Let A and B represent arbitrary
members of a population of conversants P, i an arbitrary element of sequential
indices I, $\sigma$ an arbitrary member of the set of sentences $\Sigma$.

\begin{quote}
{\sc Acceptance Inference Rule} (AIR)\\
An utterance event U $=$ (A, B, i, $\sigma$) $\leadsto_\Delta$  MS(P,
accept(B, $\cal R$($\cal S$, U)))
\end{quote}

Each assumption in $\Delta$ has an associated endorsement, which specifies the
strength of evidence supporting the assumption.  The model distinguishes between
three levels of endorsements: {\sc hypothesis, default} and {\sc linguistic},
where {\sc hypothesis} is a weaker endorsement than {\sc default}, which is
weaker than {\sc linguistic}, as represented in \ex{1}:\footnote{In earlier work,
I used five levels of endorsement and Galliers' theory of belief revision uses
many more endorsement types.  However for modelling mutual belief, as opposed to
single agent belief, most important distinctions can be made with only three
levels. These three levels could be implemented as levels of the hierarchy in
HAEL \cite{AK88}.}

\enumsentence{
{\bf Ranking on endorsement types}: \\ {\sc hypothesis} $<$ {\sc default} $<$ 
{\sc linguistic} }

These three levels are used to represent the difference between expectations and
inferred beliefs for which there is little evidence available in the shared
environment ({\sc hypothesis}), inferred mutual beliefs which are inferred as a
result of default inference rules ({\sc default}), and mutual beliefs for which
explicit linguistic evidence has been provided ({\sc linguistic}).

The initial endorsements for each
assumption above are shown below:

\begin{center}
\begin{tabular}{|lr|}
\hline &  \\
$\Delta$(U) & ENDORSEMENT \\ \hline \hline & 
\\ attend (B, U) & hypothesis \\ 
hear (B, U) & hypothesis \\ 
realize(B, U, $\cal R$($\cal
S$, U)) & hypothesis \\ 
license(B, U, $\cal L$($\cal S$, U)) &  hypothesis \\ 
accept (B, U, $\cal R$($\cal S$, U)) & hypothesis \\
\hline
\end{tabular}
\end{center}

In other words, after speaker A says an utterance, the understanding and
acceptance assumptions are licensed as hypotheses only. The basis for this
initial endorsement is that speaker A would not make the utterance unless s/he
expected the utterance could have an effect on B's mental state, but this
expectation is only a hypothesis since B has as yet provided no evidence of
mutuality.

In order to make use of these mechanisms, we must specify how the endorsements on
assumptions are combined to affect the endorsement on the consequent of an
inference rule such as the AIR. In other words, if a discourse situation
indicates P and R and P $\wedge$ R $\rightarrow$ Q, what is the endorsement on Q?
This must depend on the endorsements on P, R, and P$\wedge$ R $ \rightarrow$ Q.
The combination rule is based on the intuitive notion that a chain of reasoning
is only as strong as its weakest link:

\begin{quote}
{\sc weakest link rule}: The endorsement of a belief P
depending on a set of underlying assumptions $a_i,...a_n$
is MIN(endorsement ($a_i,...a_n$))
\end{quote}

The {\sc weakest link rule} means that for all inference rules that depend on
multiple assumptions, the endorsement of an inferred belief is the weakest of the
supporting beliefs. Thus in the case of the Acceptance Inference Rule (AIR), the
inference of acceptance is only endorsed as a {\sc hypothesis} after A's
assertion. This is a key aspect of the model that allows it to model the types of
rejection discussed above, as will be discussed in more detail below.

Speaker B's response can increase the evidence supporting these assumptions,
making them less defeasible by increasing the level of endorsement.  For example,
consider the repetition in dialogue \ref{put-that-examp}, repeated here for convenience:

\enumsentence[\ref{put-that-examp}]{
(26) H: That's right. as they come due, give me a call, about a week
in advance.  But the first one that's due the 25th, {\it let's put
that into a 2 and a half year certificate} \\ (27) E: {\bf Put that in a 2
and a half year}.  Would ... \\ (28) H: Sure. We should get over 15
percent on that.
}

According to the model the effect of the repetition is as follows:

\begin{quote}
\begin{tabular}{|lr|}
\hline &  \\
$\Delta$(U$_{26})$ & ENDORSEMENT \\ \hline \hline &  \\
 attend(E, U$_{26}$) & linguistic \\ 
 hear(E, U$_{26}$) & linguistic \\ 
 realize(E, U$_{26}$, $\cal R$($\cal S$, U$_{26}$)) & default \\
 accept(E, U$_{26}$, $\cal R$($\cal S$, U$_{26}$)) & default \\
\hline
\end{tabular}
\end{quote}

Assumptions are upgraded from an endorsement type of hypothesis to an endorsement
type of default, at a minimum, after speaker B has an opportunity to
respond.If B
responds with an explicit acceptance, with no new information such as a prompt,
repetition or paraphrase, B asserts understanding by providing evidence that
upgrades some of the assumptions $\Delta$, that speaker A made in producing the
utterance, from their initial endorsement type of hypothesis to an endorsement
type of linguistic. The less explicit forms upgrade fewer assumptions, and these
end up with an endorsement of default.  Figure \ref{ass-fig} shows how the
combination of endorsement types predicts that the different types of acceptance
IRUs have slightly different effects on the common ground.

However, note that after B's response with any of the forms under discussion, the
acceptance assumption is only endorsed as a default.  This implements the
distinction between asserting understanding and implicating acceptance discussed
in section \ref{acc-sec} \cite{Walker92a}.  Thus, this analysis accounts for the
effect of various forms of acceptance and for the effect of the {\sc
collaborative principle}.

\begin{figure*}[htb]
\begin{small}
\begin{center}
\begin{tabular}{|l|l|c|}
\hline & &\\
\thicklines
NEXT & ASSUMPTION & ENDORSEMENT \\ 
Utterance Type & ADDRESSED & TYPE\\ 
\hline \hline & &\\
PROMPT  & attention & linguistic \\ \hline & & \\
REPETITION & attention, hearing  & linguistic \\
\hline & & \\
PARAPHRASE  & attention, hearing, realize & linguistic \\
\hline & & \\
INFERENCE   & attention, hearing, realize, license & linguistic \\ 
\hline & & \\ 
ANY Next&      &  \\
Utterance  &  attention, hearing, realize, license, accept  & default \\
\hline
\end{tabular}
\caption{How different types of acceptance responses U$_{i+1}$, upgrade the endorsements
on the $\Delta$ assumptions associated with U$_i$.}
\label{ass-fig}
\end{center}
\end{small}
\end{figure*}

In addition, the default endorsement type directly implements inferences whose
consequences have the logical status of defaults, such as the scalar implicature
inference rule (SIIR), the Belief Persistence Rule and the Intention Persistence
Rule.  In other words, in addition to the AIR, the other inference rules produce
consequents that are endorsed as defaults. \footnote{The proposed analysis can be
related to default inference rules such as Perrault's by specifying several
different Belief Transfer Rules, which have different conditions on the
antecedents, and whose conclusions are annotated for level of belief.  For
example, the initial level of support for acceptance as a hypothesis can be
represented by a rule such as \ex{1} which can be defeated by rules whose
conclusions are endorsed as {\sc defaults}.

\enumsentence{ {\sc  Hypothesis Belief Transfer Rule}: \\
Say(A,B,p,t)  $\rightarrow$ Bel (B,p,t).{\sc hypothesis}
}
}

Moreover, the proposed model can straightforwardly account for the various forms
of rejection, and explain how the inference of rejection goes through. The key
modification is the replacement of the inference rules of previous accounts
\cite{Perrault90}, with the weaker Acceptance Inference Rule (AIR) which endorses
consequents as hypotheses rather than defaults, until after B's response.  The
AIR replaces both the Belief Transfer and Intention Adoption Rules of previous
accounts \cite{AK88,Perrault90}. The consequents of these two rules can then be
defeated by any of the implicature or default inference rules that can indicate
rejection.

For example, consider the case of implicature rejections, as exemplified by
\ref{new-orleans-examp}. Figure \ref{rej-imp-fig} shows how the context is
updated correctly as a result of a combination of inference rules for (1) Grice's
{\sc quality maxim}, that a speaker believes what s/he says, (2) the Scalar
Implicature Inference Rule (SIIR), (3) the Mutual Supposition Induction Schema
(MSIS), and (4) the Acceptance Inference Rule (AIR).

\begin{figure*}[htb]
\begin{small}
\begin{center}
\begin{tabular}{|l|c|c|l|}
\hline & & &\\
Utterance & 		Mutual Suppositions 		&Endorsement 	&Rule\\ 
          &  OF P && \\ \hline \hline & & & \\
A: We bought these pajamas & Bel(A, $\cal R$($\cal S$, U$_{1}$)) 		& linguistic 	& Quality\\ 
 in New Orleans for me.   & $\cal R$($\cal S$, U$_{1}$)			& hypoth 	& AIR\\ \hline & & & \\

B: We bought these pajamas 		& Bel(B, $\cal R$($\cal S$, U$_{2}$)) 		& linguistic 	& Quality\\
in $_{FR}$New Orleans &$\neg$Bel(B,$\cal R$($\cal S$, U$_{1}$))
& default 	& SIIR, (P, P $\wedge$ Q)\\
 					& $\neg$$\cal R$($\cal S$,
U$_{1}$)		& default	& MSIS and previous\\
					& $\cal R$($\cal S$, U$_{2}$)			& hypoth  	& AIR \\ \hline 
\end{tabular}
\end{center}
\end{small}
\caption{Effect of Rejection Implicatures}
\label{rej-imp-fig}
\end{figure*}

As figure \ref{rej-imp-fig} shows, A's assertion provides linguistic evidence of
A's belief in the content of the assertion by the Quality Maxim. The content is
added to the common ground as a hypothesis only, on the basis of the AIR. B's
response first adds that B believes the content of B's assertion to the common
ground with an endorsement of linguistic by the Quality Maxim. Then the SIIR with
the scale $\cal O$ of conjunctive assertions leads to the inference of the
implicature rejection. The consequent of the SIIR is a default, so the
implicature is added to the common ground as a default. On the basis of this
addition, the content of the implicature is then added to the common ground as a
default using the MSIS.  As figure \ref{rej-imp-fig} shows, the result is that
the consequent of the SIIR, which is endorsed as a default, defeats the speaker's
original belief about acceptance, which is only endorsed as a hypothesis.
Finally, the fact that B's utterance is a partial acceptance is reflected by the
fact that the proposition that B's utterance realizes is added to the common
ground as a hypothesis, by means of the AIR.\footnote{Section \ref{disc-sec}
discusses whether the content of B's utterance should be stronger than a
hypothesis.}

The same mechanism accounts for epistemic rejections such as asserting an
inconsistent past belief. The Belief Persistence Rule is a default rule with no
underlying assumptions. Thus the assertion of an inconsistent past belief leads
to the addition of an inconsistent current belief endorsed as a {\sc
default}. Since the consequent of the AIR is only endorsed as a hypothesis until
after B has taken a turn, the default of the inconsistent past belief defeats the
inference of acceptance, in the same way that the SIIR default defeats the inference
of acceptance in the example in figure \ref{rej-imp-fig}.

Finally, the treatment is identical for rules for inferring the adoption of
intentions. The Intention Persistence Rule is a default rule with no underlying
assumptions. The Intention Adoption Rule is subsumed under the AIR simply by
positing a different process behind B's decision to accept (deliberation for
intentions rather than belief revision for beliefs). So if speaker A makes a
proposal, the inference of intention adoption is only endorsed as a hypothesis
until after B's turn. Thus the conflicting defaults mentioned in the discussion
of example \ref{consider-cash-examp} are resolved in the same manner: the
consequent of the Intention Persistence Rule is a default belief, and this
default defeats the consequent of the AIR which has replaced the Intention
Adoption Rule.

\section{Discussion of Open Issues and Future Work}
\label{disc-sec}

Above we claimed that the partial acceptance nature of implicature rejections
is modelled by the model of the common ground in which B's utterance is added
to the common ground as a hypothesis. This hypothesis will be upgraded to a
default after A's next turn, if A does not reject B's utterance. Since A has
previously asserted the content of B's utterance, we can assume this upgrade
would occur. An open issue is whether this is adequate for representing the
nature of partial acceptance. 

There are two possible ways to address this issue. The first way requires a
revision of Hirschberg's theory so that the SIIR would only add to the context
the implicature, rather than the negation of all of A's utterance. In other
words, in figure \ref{rej-imp-fig}, the implicature listed as $\neg$Bel(B,$\cal
R$($\cal S$, U$_{1}$)) would be replaced by an implicature for only a
subproposition of $\cal R$($\cal S$, U$_{1}$), informally $\neg$Bel(B, ``for
you'').  This would also require that the SIIR was sensitive to the type of scale
$\cal O$, in particular it would have to treat the conjunctive scale as different
from other scales. The ramifications of these modifications to Hirschberg's
theory are beyond the scope of this paper.

A second way of addressing this issue would be that the update rules for the
common ground would be more intelligent and note that A had already committed
himself to an assertion that entails what B asserted. On the basis of these
facts, B's assertion would have a stronger endorsement than hypothesis to start.
Again, the ramifications of these modifications are beyond the scope of this
paper.

Another type of issue has to do with the potential additional functions of the
various forms of acceptance. The analysis here would suggest that the {\bf
function} of these forms is to increase the endorsements on assumptions in the
common ground. However, although these forms {\bf can} be used to address the set
of underlying assumptions, it is not at all clear that this is
{\bf why} a conversant chooses one form over another. For example, repetitions
are often used in the financial advice talk show, but the broadcast is very
clear, and there are no examples where the caller misheard the talk show
host. There seems to be no reason why the callers would feel it is necessary to
provide evidence of verbatim hearing.

In addition, repetitions and paraphrases are usable as a prompt to the other
speaker to say more about a topic. These utterances seem systematically ambiguous
between backchannels and prompts. For example, in dialogue \ref{put-that-examp},
E after producing the repetition in 27, attempts to continue her turn by asking
for more information. H interrupts her as though she had been prompting him for
more support for his advice.

The final issue has to do with generalizations of the account presented here.
Implicatures can be used in indicating acceptance, and in other types of
rejection.  Consider \ex{1} (Hirschberg's (106) p. 90):

\eenumsentence{
\item[A:]  A very large and vicious dog is about to attack me.
\item[B:]  He's large.
\label{dog-examp}
} 

B's utterance was less informative than it might have been, leading to the
implicature in \ex{1}:

\enumsentence{He's not vicious.}

It would also have been possible for B to respond as in \ex{1}B:

\eenumsentence{
\item[A:]  A very large and vicious dog is about to attack me.
\item[B:]  He's not large.
\label{dog-examp-denial}
} 

Rather than generating an implicature rejection, B's utterance is an explicit
rejection, which however implicates partial acceptance.  According to
Hirschberg's theory, the denial in \ex{0}B leads to the implicature in \ex{1}, by
which the speaker implicates acceptance of the part of A's utterance that was not
rejected.

\enumsentence{A vicious dog is about to attack you.}

This shows that basing the account here on Hirschberg's theory makes it very
general, once we have a mechanism for adding to the common ground that lets
implicatures be licensed in the context of an assertion. While there were no
naturally occurring examples of rejecting proposals by implicature, the
conditions on rejection by implicature will also cover proposals. For example, it
is possible to reject a proposal with an {\sc alternate} scalar item as in \ex{1}B:

\eenumsentence{
\item[A:] Let's buy some bananas.
\item[B:] Let's buy some oranges.
}

It is also possible to reject an assertion with a {\sc higher} scalar item as in
\ex{1}B \cite{Horn89}:.\footnote{This is a variation on the metalinguistic
negation in \ex{0}, in which the speaker is apparently inconsistent:

\enumsentence{
`Still', Edwin concludes, `I did rather like him, didn't you?' \\
`No', Vinnie says... `I didn't ``rather like Chuck'', if you want
to know. I loved him'. (Horn's 75', p 401, from Lurie's {\it Foreign
Affairs}, p. 420)
}}

\eenumsentence{
\item[A:] Vinnie likes Chuck.
\item[B:] She loves him.
}

The existence of these generalizations highlights a final limitation of the current study: the corpus on which
this analysis was based is small, and the classification of utterances into acceptances and
rejections was based on the author's interpretations. Filling in the analysis here and generalizing it
must be based on a larger corpus for which speech is available, so that we can determine whether
the classification proposed can be reliably applied in a larger corpus. 

\section{Conclusion}

This paper discussed the processes by which conversants in a dialogue can infer
whether their assertions and proposals have been rejected by their conversational
partners, and specified cues that can be used in this process. This discussion
expanded on previous work by showing that logical consistency is a necessary
indicator of acceptance, but that it is not sufficient, and that logical
inconsistency is sufficient as an indicator of rejection, but it is not
necessary. 

Section \ref{acc-sec} discussed all the various forms of logically consistent acceptances,
such as repetitions, paraphrases and utterances that make inferences explicit.  I showed
by a distributional analysis that the prosodic realization of these utterances marks them as 
old and predictable information and suggested that hearers can use these prosodic markers
as cues for interpretation. 

In section \ref{reject-sec}, I analyze the different types of rejection found in a
small corpus from the financial advice domain. I build on Horn's analysis which
suggests that the types of rejection include: (a) {\sc denial}; (b) {\sc logical
contradiction}; (c) {\sc implicit denial}, where B denies a presupposition of
A's; and (d) {\sc refusal}, also called {\sc rejection} where B refuses an offer
or proposal of A's \cite{Horn89}.  The corpus analysis highlights the existence
of three new classes of rejection: {\sc implicature rejections}, {\sc epistemic
rejections} and {\sc deliberation rejections}. I show how these rejections are
inferred as a result of default inferences, which, by other analyses, would have
been blocked by the context.

In section \ref{focus-sec}, I discuss how conversants can use information
structure and prosody as well as logical reasoning in distinguishing between
acceptances and logically consistent rejections. Acceptances re-realize focal
information from U$_1$ and mark it as old information.  Rejections re-realize the
open proposition from U$_1$, and replace the focal item with a scalarly related
item. I proposed that the {\sc substitution of focus} condition can be used in
dialogue to determine when {\sc implicature rejections} are licensed, and discuss
the use of Fall-Rise intonation as an additional cue for rejection
\cite{Ladd80,Horn89,Hirschberg85}.

Finally, I argue that these observations show that we need a model of the common
ground that models the difference between implicit and explicit acceptance,
explains how implicatures can be used for indicating rejection, and explains how
conflicting defaults are resolved in the inference of rejection. In section
\ref{lewis-sec}, I propose a model of the common ground that allows these default
inferences to go through, and show how this model, which was proposed to account
for the wide variety of indications of acceptance, also models the new types of
rejection.

\section{Acknowledgements}

I am indebted to Megan Moser for originally suggesting that example
\ref{garage-examp} might be a type of scalar implicature, and to
Beatrice Santorini, Nick Asher, Craige Roberts, Masayo Iida, Beth Ann
Hockey, Ellen Prince, Rich Thomason, Larry Horn, Jon Oberlander, Steve
Whittaker and two anonymous reviewers for discussion and critical
comments.


\end{document}